\documentclass[runningheads]{llncs}

\usepackage{eccv}

\usepackage{framed}

\usepackage{eccvabbrv}

\usepackage{wrapfig}
\usepackage{comment}
\usepackage{graphicx}
\usepackage{booktabs}
\usepackage[dvipsnames]{xcolor} 
\usepackage{listings} 
\definecolor{commentcolor}{rgb}{0.5,0.5,0.5} 
\definecolor{keywordcolor}{rgb}{0,0,1} 
\definecolor{stringcolor}{rgb}{0.58,0,0.82} 

\usepackage[accsupp]{axessibility}  

\usepackage{hyperref}

\usepackage{orcidlink}

\usepackage{multirow}

\usepackage{multicol}
\usepackage{algorithm}
\usepackage{algpseudocode}
\usepackage{lipsum}
\usepackage{pgfplots}

\usepackage{arydshln}

\usepackage{pifont}

\newcommand{\cmark}{\ding{51}}%
\newcommand{\xmark}{\ding{55}}%

\usepackage[most]{tcolorbox}
\definecolor{block-gray}{gray}{0.95}
\newtcolorbox{myquote}{colback=block-gray,grow to right by=-0mm,grow to left by=-0mm,
boxrule=0pt,boxsep=0pt,breakable}
\makeatletter
\def\quoteparse{\@ifnextchar`{\quotex}{\singlequote}}
\def\quotex#1{\@ifnextchar`{\triplequote\@gobble}{\doublequote}}
\makeatother
\def\singlequote#1`{[StartQ]#1[EndQ]\quoteON}
\def\doublequote#1``{[StartQQ]#1[EndQQ]\quoteON}
\long\def\triplequote#1```{\begin{myquote}\parskip 1ex#1\end{myquote}\quoteON}
\def\quoteON{\catcode``=\active}
\def\quoteOFF{\catcode``=12}
\quoteON
\def`{\quoteOFF \quoteparse}
\quoteOFF

\begin{document}

\title{iHuman: Instant Animatable Digital Humans From Monocular Videos}

\titlerunning{iHuman: Instant Digital Humans}

\author{Pramish Paudel\inst{1}\orcidlink{0009-0006-3145-5077} \and
Anubhav Khanal\inst{1}\orcidlink{0009-0003-7787-6414} \and
Ajad Chhatkuli \inst{2,3}\orcidlink{0000-0003-2051-2209} \and
Danda Pani Paudel \inst{2,3,4}\orcidlink{0000-0002-1739-1867} \and 
Jyoti Tandukar \inst{1}}
\authorrunning{P. Paudel et al.}

\institute{Pulchowk Campus, IOE, Tribhuvan University, Lalitpur, Nepal \\ \and
ETH Zürich, Zürich, Switzerland \\ \and
NAAMI, Kathmandu, Nepal \\ \and
INSAIT, Sofia, Bulgaria
}

\maketitle
\begin{figure}
    \centering
    \includegraphics[width=0.85\linewidth]{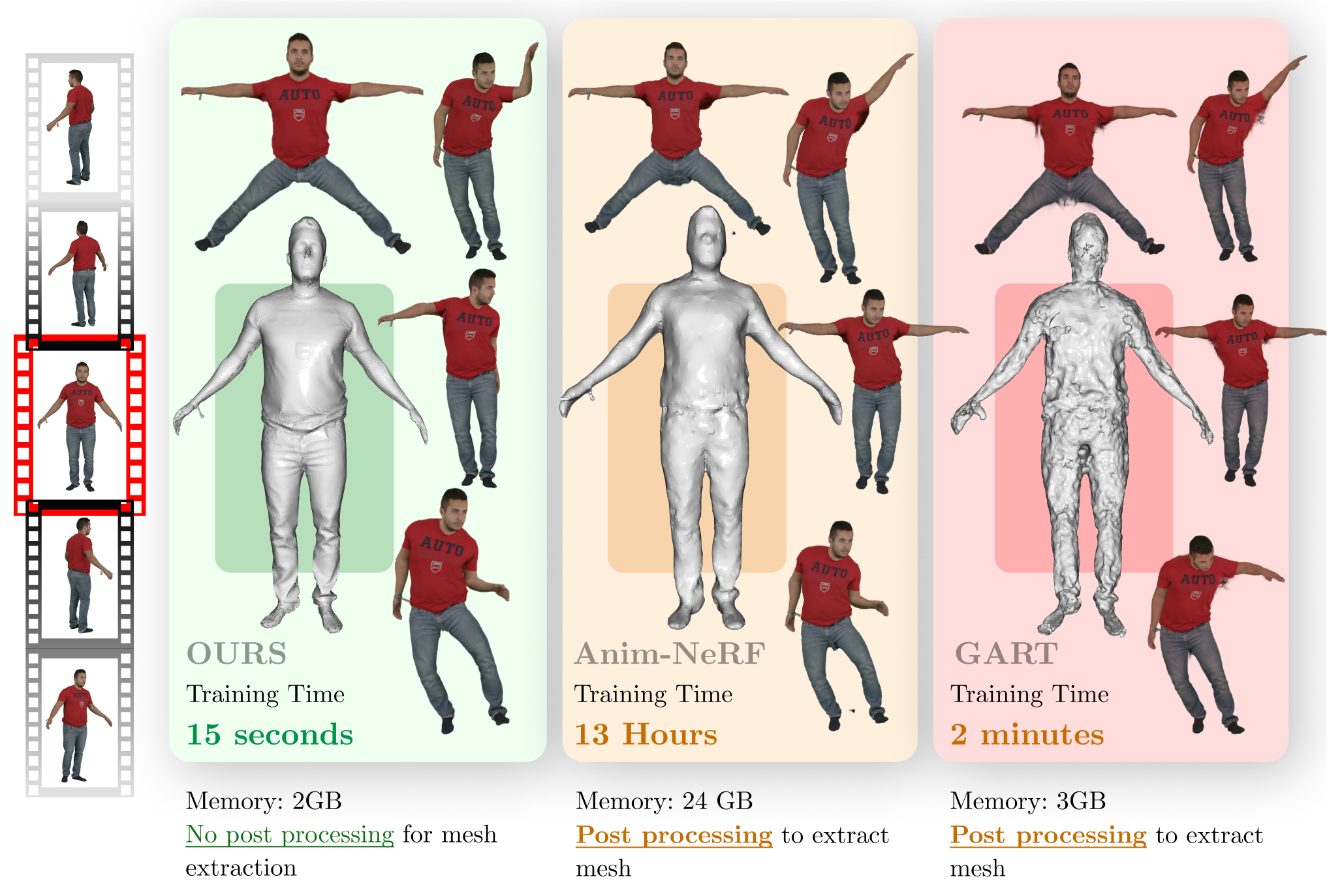}
    \vspace{-3mm}
    \caption{\textbf{Instant Digital Humans}. Our method provides detailed and accurate 3D mesh and renderable Gaussian Splats instantly in 15 seconds of training time, from a  monocular video. In contrast, the existing methods Anim-NeRF~\cite{DBLP:conf/cvpr/PengZXWSBZ21} and GART~\cite{lei_gart_2023} provide lower quality mesh and rendered images, even after using  more training time and compute. Input video (left) and  rendered poses around the recovered meshes.}
    \vspace{-10mm}
    \label{fig:teaserFigureQalitative}
\end{figure}

\begin{abstract}
Personalized 3D avatars require an animatable representation of digital humans. Doing so instantly from monocular videos offers scalability to broad class of users and wide-scale applications. In this paper, we present a fast, simple, yet effective method for creating animatable 3D digital humans  from monocular videos. Our method utilizes the efficiency of Gaussian splatting to model both 3D geometry and appearance. However, we observed that naively optimizing Gaussian splats results in inaccurate geometry, thereby leading to poor animations.  


This work achieves and illustrates the need of accurate 3D mesh-type modelling of the human body for animatable digitization through Gaussian splats. This is achieved by developing a novel pipeline that benefits from three key aspects: (a) implicit modelling of surface's displacements and the color's spherical harmonics; (b) binding of 3D Gaussians to the respective triangular faces of the body template; (c) a novel technique to render normals followed by their auxiliary supervision. Our exhaustive experiments on three different benchmark datasets demonstrates the state-of-the-art results of our method, in limited time settings. In fact, our method is faster by an order of magnitude (in terms of training time) than its closest competitor. At the same time, we achieve superior rendering and 3D reconstruction performance under the change of poses. Our source code will be made publicly available. 


  \keywords{Digital Humans\and Gaussian Splats \and Surface reconstruction}
\end{abstract}

\section{Introduction}
\label{sec:intro}

Instant and accurate creation of personalized 3D avatars is highly sought-after for digital human representation, to enable vast applications in virtual reality (VR), augmented reality (AR), gaming, and telepresence. A key component in this regard is the animatable representation~\cite{huang2020arch,peng2021animatable,feng2021learning}. On the other hand, reconstructing animatable digital humans instantly from monocular videos can immediately facilitate wide-scale applications serving a broad class users. 
The most existing monocular video based methods focus on either the real-time rendering solutions~\cite{alldieck2018detailed} (using long training/reconstruction time), or only mesh-level reconstruction~\cite{goel2023humans} without the possibility of realistic re-rendering under the change in pose. These solutions eventually hinder the broad-scale applicability, which we aim to address in this paper by developing a novel method for instant and accurate modelling of animatable digital humans from monocular videos. 

\begin{wrapfigure}[10]{r}{0.5\linewidth}
  \begin{center}
  \vspace{-33pt}
    \includegraphics[width=0.48\textwidth]
{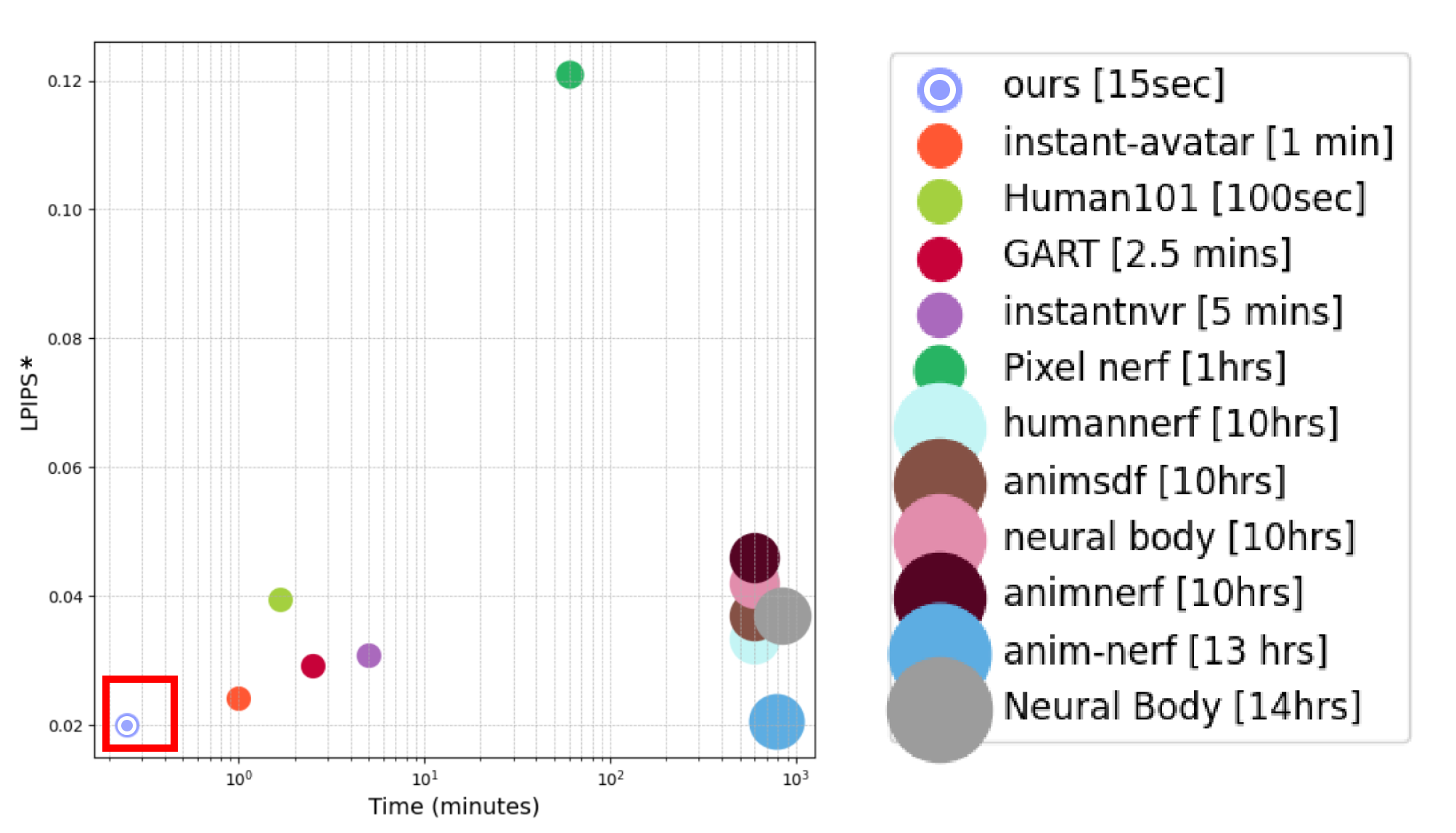}
      \end{center}
      \captionsetup{aboveskip=-4mm, belowskip=0cm}
    \caption{Training time (mins $\downarrow$) vs. rendering ($\downarrow$) comparison for different methods. }
    \label{fig:teaserFigQuant}
\end{wrapfigure}

In recent year, building on the remarkable success of representing radiance fields implicitly (i.e. NeRF~\cite{mildenhall2020nerf}), several methods~\cite{chen2021animatable,chen2023fast,chen2021snarf, guo2023vid2avatar, jiang2022instantavatar, DBLP:conf/eccv/JiangYSTR22, He_2021_ICCV, marc2023hd, geng2023learning, wang2022neural, Gafni_2021_CVPR, He_2021_ICCV, jiang2022selfrecon, zhang2021editable, kwon2021neural, kwon2021neural, 9710330, DBLP:conf/cvpr/PengZXWSBZ21, saito2020pifuhd, DBLP:conf/eccv/LiTVZGKL22, liu2021neural, Noguchi_2021_ICCV, DBLP:conf/eccv/WangSGT22, 9879404, xiu2021icon, su2022danbo, xu2021hnerf, su2021anerf, yu2023monohuman, DBLP:conf/cvpr/ZielonkaBT23, Zheng_2022_CVPR}, have been developed to capture high-fidelity humans from multiple frames of videos. However, the high computational demand of the volume rendering in the NeRF-based methods~\cite{zhu2023trihuman,yariv2021volume,shahbazi2023nerf} creates a major bottleneck for the aimed instant animatable digitization. Therefore,  some  very recent methods~\cite{lei_gart_2023, li_human101_2023,li2024gaussianbody,yuan2023gavatar}\ have been developed by leveraging the rendering efficiency of the Gaussian Splats~\cite{kerbl20233d}. However, these methods do not meet some or all of the required criteria in capturing (i) from monocular videos; (ii) in instant manner; (iii)  animatable avatar; and (iv) high quality re-rendering under change in pose.

In this paper, we propose an efficient pipeline to convert a monocular video, with known pose, to animatable digital humans instantly (training time on par with that of capture) in few seconds, using Gaussian splats based modelling.  Doing so faithfully is particularly challenging primarily due to the difficulty of  (i) inaccurate initialization of the Gaussian splats and (ii) ensuring the animatable nature of the output. Our proposed method addresses the challenges by introducing  contributions on three aspects: (a) implicit modelling of surface's displacements and the color's spherical harmonics; (b) binding of 3D Gaussians to the respective triangular faces of the body template; (c) a novel technique to render normals followed by their auxiliary supervision. 
The proposed method is intuitive, simple, fast, yet effective. The qualitative and quantitative benifits of our method are highlighted in Figure~\ref{fig:teaserFigureQalitative} \&~\ref{fig:teaserFigQuant}, respectively. 

The main contributions of this paper are:\\
-- We propose a complete pipeline to obtain accurate 3D mesh bound Gaussian splats suitable for avatar animation, from monocular videos in instant manner. \\
-- The proposed technical contributions involve; implicit representation, binding of gaussians to triangular faces, normal derivation for the auxiliary supervision.\\
-- We conduct exhaustive experiments for comparisons, where our method achieves superior representation quality with an order of magnitude faster training speed. \\

\section{Related Work}
\label{sec:related work}

\paragraph{Mesh based reconstruction methods.}
Most methods that represent the human body as a mesh make use of SMPL~\cite{matthew2015smpl} or other parametric body models~\cite{dragomir2005scape, DBLP:conf/eccv/OsmanBB20, Pavlakos_2019_CVPR}.
Methods in this category predict the parameters for the parametric body model either by regression~\cite{pavlakos2018learning, kanazawa2017endtoend, omran2018neural} or by optimization~\cite{Bogo:ECCV:2016}. Kolotouros et al.~\cite{kolotouros2019convolutional} and similar methods~\cite{moon2020i2lmeshnet, lin2021mesh} directly regress the 3D vertices. Although the output meshes here can be animated they do not contain the clothing details and personalized facial features. Methods which extend the parametric body model with a deformation layer \cite{DBLP:conf/cvpr/AlldieckMBTP19, DBLP:conf/cvpr/AlldieckMXTP18, DBLP:conf/eccv/BhatnagarSTP20, DBLP:conf/nips/BhatnagarSTP20} can model clothing as well but are unable to accurately model personalized geometric details.
\paragraph{Implicit functions based approaches.}
Implicit functions based reconstruction methods \cite{mildenhall2020nerf, DBLP:conf/cvpr/ParkFSNL19, mescheder2018occupancy, NEURIPS2021_25e2a30f, wang2021neus} use an MLP to learn an implicit function such as occupancy, signed distance fields or density fields to describe geometry. They can represent and render the geometric details of static scenes but suffer from high training time. Anim-NeRF\cite{chen2021animatable} and other similar methods \cite{DBLP:conf/cvpr/PengZXWSBZ21, peng2021animatable, liu2021neural, wang2022neural, 9879404, huang2020arch, He_2021_ICCV,DBLP:conf/eccv/JiangYSTR22,DBLP:conf/eccv/LiTVZGKL22, DBLP:conf/eccv/WangSGT22, chen2021snarf, chen2023fast, guo2023vid2avatar, jiang2022selfrecon} extend NERF to dynamic scenes by using SMPL~\cite{matthew2015smpl} guided deformations between the observed space and a static canonical space allowing for explicit control. Instant Avatar \cite{jiang2022instantavatar} and similar approaches \cite{jiang2023instantnvr, DBLP:journals/tog/ZhaoJYZWDZZWXY22} use \cite{thomas2022instant} to speed up the training time but still have high memory requirements.
\paragraph{Gaussian Splat based approaches.}
Recently introduced 3D gaussian Splatting(3D-GS)~\cite{DBLP:journals/tog/KerblKLD23} uses 3D Gaussians and its projections to represent a static scene. 3D-GS achieves significantly faster training and rendering time over NERF-based approaches. Recent works~\cite{li_human101_2023, pang_ash_2023, lei_gart_2023, moreau_human_2023, qian_3dgs-avatar_2023, kocabas_hugs_2023, li_animatable_2023, jena_splatarmor_2023, zielonka_drivable_2023} extend 3D-GS to represent dynamic scenes using SMPL guided deformations. They produce an animatable representation of the human body at speed compared to previous methods. However, the Gaussians obtained from the standard 3D-GS optimization process are unstructured and may not correspond to the surface, thus the Poission's reconstruction~\cite{kazhdan2006poisson} does not result in accurate geometry. A 3D-GS extension, SuGaR~\cite{guédon2023sugar} introduces an approach to align Gaussians to the surface geometry. Consequently, a mesh can be extracted using Poission's reconstruction~\cite{kazhdan2006poisson}. Other methods \cite{waczynska2024games}, \cite{qian2023gaussianavatars} use a mesh prior to initialize the 3D Gaussians to obtain well structured 3D Gaussians. Binding Gaussians to the mesh surface produces better geometry than approaches that only use 3D Gaussians but they still do not capture surface details. \cite{saito2020pifuhd, xiu2021icon, xiu2022econ} use normal maps to capture high frequency details. However, \cite{saito2020pifuhd, xiu2021icon, xiu2022econ} uses a single image to infer geometry, thereby resulting in relatively less accurate geometry. Our method uses Gaussians that are binded to the surface of a mesh as well as a novel normal guidance to produce animatable mesh of the human body while capturing  body details. 
\section{Method}
\label{sec:method}
Provided a monocular video sequence with a dynamic human and the body poses, our goal is to generate a personalized colored mesh 3D model of a subject consisting of body shape, hair and clothing geometry, and underlying skeleton. Given an $n$ frame video sequence $(I_t)_{t=1}^n$ of a single subject in front of a fixed camera (camera pose and intrinsics), along with the respective body poses $\{\theta_t\}$ we output a personalized animatable representation of the human subject. The keyword `animatable' implies that we should be able to render the underlying representations in novel body poses $\{\theta_j\}$. Additionally, we want to complete the challenging training process in seconds, in favor of scalability.  
    
\begin{figure}[h]
    \centering
    \includegraphics[width=1\linewidth]{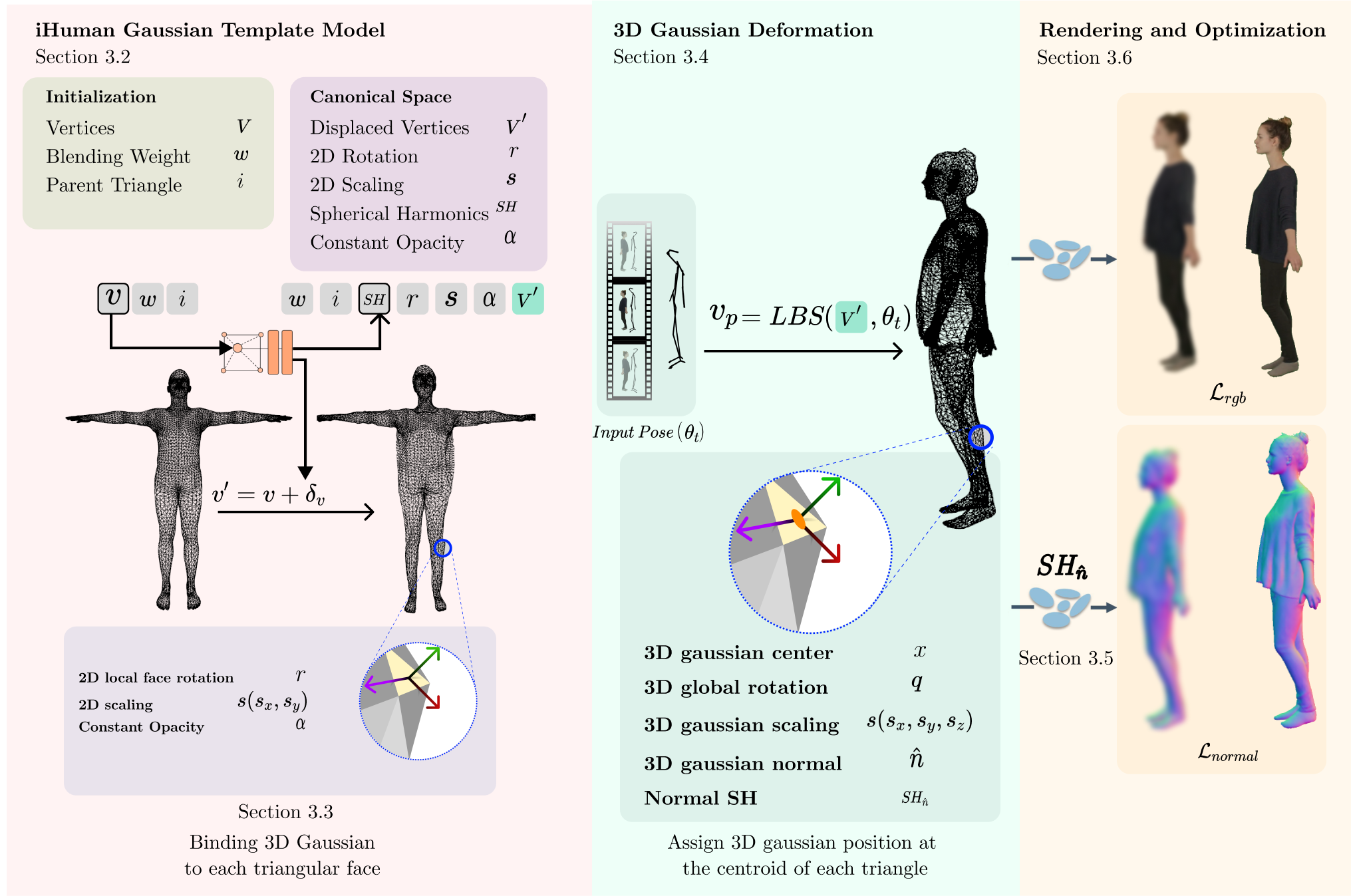}
    \caption{Our method represents the human body in canonical space with gaussians parameterized by 3D gaussian centers $x$, rotations $q$, scales $S$, opacity $\alpha_o$, colors $SH$, skinning weight $w$ and its associated parent triangle $i_x$. It takes body pose $\theta_t$ of $t^{th}$ frame as input and applies forward linear blend skinning to transform $v'$ to posed space $v_p$. We compute gaussian center $x$ from the posed space vertices $v_p$ of $i_x$. The normal of parent triangle $i_x$ is encoded to $SH_{\hat{n}}$ and rasterized to obtain the normal map $I_{\hat{n}}$. Then, we apply photometric loss and normal map loss to recover both geometry and color. The GT normal map ($\bar{I}_{\hat{n}}$) is obtained from monocular RGB image ($I_n$) using pix2pixHD \cite{wang2018high} network.}
    \label{fig:method}
    \vspace{-5mm}
\end{figure}

We achieve our goal of obtaining an animatable 3D human using 3D Gaussian Splatting (3D-GS)~\cite{kerbl20233d,guédon2023sugar}. Below we explain the 3D-GS and its deformations as preliminaries in \S \ref{sec:preliminaries}. We then introduce our iHuman representation and describe the details of our method. iHuman consists of initializing 3D-GS in the canonical SMPL pose, see \S \ref{sec:animatable_3d_gaussian}. We bind each 3D Gaussian to a triangle face as described in \S \ref{sec:binding_gaussians_to_mesh_surface}. We then proceed onto deforming the 3D-GS consistent to the posed space, corresponding to the real image in \S \ref{sec:3d_gaussian_deformation}. Taking advantage of explicit 2D Gaussians embedded in 3D~\cite{guédon2023sugar,waczynska2024games}, we encode normal for each Gaussian in \S \ref{sec:normal_map_from_3d_gaussian}.


\subsection{Preliminaries}
\label{sec:preliminaries}
\textbf{3D Gaussian Splatting.}
3D Gaussian Splatting (3D-GS) has recently become the state-of-the-art tool for novel view synthesis. Important for our application, different from NERF, 3D-GS also uses explicit 3D representation using anisotropic 3D Gaussians.

A 3D Gaussian can be written in terms of its full 3D covariance matrix $\Sigma \in \mathbb{R}^{3\times3}, \Sigma \succeq 0$ and position in space $y \in \mathbb{R}^3$ along with center $\mathsf{x} \in \mathbb{R}^3$.
\begin{equation}
\label{eq:std_gaussian}
G(y) = \exp \left(-\frac{1}{2}(y-x)^\top\Sigma^{-1}(y-x)\right).
\end{equation}

Kerbl et al.,~\cite{DBLP:journals/tog/KerblKLD23} however represents each 3D Gaussian to be splatted by the Gaussian 3D center position $x \in \mathbb{R}^3$, color $c \in \mathbb{R}^3$, opacity $\alpha \in \mathbb{R}$, its orientation parametrized by a 3D rotation written as a quaternion $q \in \mathbb{R}^4$, and anisotropic 3D scaling factor $s \in \mathbb{R}^3$~\cite{DBLP:journals/tog/KerblKLD23}. Instead of directly assigning a color value $c$, we use the Spherical Harmonics function, denoted as $SH$ to model appearance on the projection ray originating from $x$. Thus the representation can be denoted as,
\begin{equation}
\label{eq:original_gaussian_representation}
\mathcal{G} = \{x, q, S, \alpha_0, SH\}.
\end{equation}

Note that Eq.~\eqref{eq:original_gaussian_representation} can be used with compact sets by approximating Gaussians with ellipsoids, thus simplifying both rendering and optimization \cite{DBLP:journals/tog/KerblKLD23}.
Nonetheless, equivalence between the two exists where $\Sigma$ can be written explicitly in terms of the scale $s$ and orientation $R(q)\in SO_3$ as
\begin{equation}
\Sigma = RS S^\top R^\top.
\end{equation}

In order to project the 3D Gaussians onto the camera for rendering, given a viewing transformation $W$, the covariance matrix $\Sigma'$ in camera coordinates is given as follows:
\begin{equation}
\Sigma' = \mathsf{J} W \Sigma W^\top \mathsf{J}^\top,
\end{equation}
where, the Jacobian $\mathsf{J}$ is the affine approximation of the projective transformation.

Finally, the projected Gaussians are depth sorted and rasterized. During the process of rasterization, each Gaussian contributes to the pixel color through an opacity value $\alpha$ and coefficients $\mathbf{c}$ which encapsulate the color information. The resulting volumetric rendering equation that each Gaussian adds to the pixel is represented as follows:

\begin{equation}
C = \sum_{i \in N} c_i \alpha_i,
\end{equation}
where, $C$ is the final pixel color and $N$ is the number of Gaussians (ellipsoids) projected on to the pixel.
\\

\subsection{Gaussian Human Template Model}
\label{sec:animatable_3d_gaussian}
Our iHuman approach uses a Gaussian template model on the canonical pose of the standard SMPL shape~\cite{matthew2015smpl}. We denote this canonical mesh as $\mathcal{M}$ composed of vertices $V_{\mathcal{C}} = \{v_0, v_1, ..., v_m\}$ and triangles $F= \{i_x\}$, thus $\mathcal{M}= (V_{\mathcal{C}}, F)$. We then bind the Gaussians in 3D to the canonical mesh $\mathcal{M}$ as described in Eq.~\eqref{eq:original_gaussian_representation}. This process is described in \S\ref{sec:binding_gaussians_to_mesh_surface}, where each Gaussian is tied to a specific face, \ie, triangle $i_x$. We thus obtain the Gaussian Splat representation for the subject in the canonical SMPL pose by extending Eq.~\eqref{eq:original_gaussian_representation} as follows:
\begin{equation}
\label{eq:skinned_gaussian}
\mathcal{G}_{\text{skinned}} = \{x, q, S, \alpha, SH, w, \delta_x, i_x\}.
\end{equation}
In contrast to Eq.~\eqref{eq:original_gaussian_representation}, in Eq.~\eqref{eq:skinned_gaussian}, each Gaussian center $x$ is in fact the centroid of a triangle $i_x$. Additionally, we introduce new parameters where $w$ is the skinning weights obtained from the standard parametric body model~\cite{matthew2015smpl}. Importantly, $\delta_v$ is the vertex displacement for the canonical shape vertex $v$ to the clothed subject shape $v'$:
\begin{equation}
\label{eq:v'}
v' = v + \delta_v.
\end{equation}
The displacement vectors $\delta_v$ are obtained vertex-wise from a continuous Hash Encoder whose output is fed to a 3-layer MLP (multi-linear perceptron). We denote this as:
\begin{equation}
\label{eq:delta_x}
    \delta_v = f_\delta\left(h(v)\right)
\end{equation}
where $f$ is a 3-layer MLP and $h(.)$ is the hash encoder similar to the instant NGP~\cite{thomas2022instant}. 

After the displacements, we bind the 3D Gaussians to the parent triangle $i_x$ by centering it on the centroid of the face $i_x$ at $x$. The rotation $q$ and scale $S$ are 2D rotation and scale as explained in \ref{sec:binding_gaussians_to_mesh_surface}.
Each triangle center $x$ for $i_x$ is obtained as, 
\begin{equation}
\label{eq:gaussian_center_from_verts}
    x=\frac{{i_x[1] + i_x[2] + i_x[3]}}{3}.
\end{equation}

In order to obtain the new Gaussian representation of Eq.~\eqref{eq:skinned_gaussian}, we need to obtain the set of parameters $\{q, S, \alpha, SH, \delta_v\}$. Using Figure~\ref{fig:method}, we now explain how we obtain these quantities.

For each mesh face $i_x$, rotation $q$, scale values $S$ and Gaussian opacity $\alpha$ are optimized as free variables. We further use optimization for the rotation $q$ and $SH$ coefficients for appearance rendering. 
Similarly, the SH coefficients are optimized as a function of $h(v)$.








\subsection{Binding Gaussians To Mesh Surface}
\label{sec:binding_gaussians_to_mesh_surface}
In the original works, the Gaussian splats are initialized on the point clouds output by a Structure-from-Motion (SfM) method such as COLMAP~\cite{schonberger2016structure}. A recent work SuGaR~\cite{guédon2023sugar} proposes using surface aligned Gaussians for the optimization, such that one of the axes of the Gaussian covariance $\Sigma$ is aligned to the surface normal $n_i$, with the corresponding scale as 0. In iHuman, we have the advantage of mesh initialization through the SMPL canonical model. Therefore, we propose to align all Gaussians according to the following steps:
\begin{enumerate}
    \item Compute the surface normal $n_i$ for each face $i_x$.
    \item For each Gaussian, assign $n_i$ as one of the directions of its covariance matrix $\Sigma$, with the corresponding scale in $S$, \ie, $S_3=\epsilon$. In practice we keep $\epsilon=1mm$, an extremely small value.
    \item Assign the other two directions according to the major directions of the triangular face.
\end{enumerate}

As a consequence of having a 2D Gaussian, we further reduce the learnable parameters required for obtaining the posed as well as canonical human mesh. For each Gaussian we can directly set $S_3=\epsilon$ and the quaternion is reduced to a complex number of a single degree of freedom in order to keep the Gaussian aligned with the triangle.



\subsection{3D Gaussian Deformation}
\label{sec:3d_gaussian_deformation}
Together with the Gaussian binding and the template model described in \S \ref{sec:animatable_3d_gaussian}, \S \ref{sec:binding_gaussians_to_mesh_surface}, we are able to precisely represent a human surface in the canonical pose. In this section, we deform the Gaussian Splat model in order to represent any pose of a human subject. Given the input pose $\theta_t$, we achieve the deformation using forward linear blend skinning~\cite{lewis2023pose}.

Thus, we compute the transformation of each vertex $v'$ in posed space with blend skinning $\omega(\theta_t)$. The transformation of each point $v$ is calculated with blend skinning $\omega(\theta)$ and target bone transformation $B(\theta_t) = \{B_1(\theta_t), \ldots, B_{n_b}(\theta_t)\}$. The skinning weight field is defined as:
\begin{equation}
w(v_c) = \{w_1, \ldots, w_{n_b}\}
\end{equation}
where $v'$ is a point in canonical space and $n_b$ is the number of bones.

Target bone transformations $B_t = \{ B_t^1, ..., B_t^{n_b} \}$ in frame $t$ can be calculated from the input poses and the corresponding skeleton as follows:
\begin{equation}
S_t, J, T_t \mapsto B_t,
\end{equation}
where $S_t = \{ \omega_t^1, ..., \omega_t^{n_b} \}$ refers to the rotation Euler Angle of each joint in frame $t$ (world rotation for $\omega_t^1$ and local rotation for the rest), $T_t$ is the world translation in frame $t$, and $J = \{ J_1, ..., J_{n_b} \}$ is the local position of each joint in canonical space. We transform a vertex in canonical space to posed space:
\begin{equation}
v_p = \sum_{i=1}^{n_b} w_i \cdot B_i^t \cdot v'
\end{equation}
where $v_p$ represents the direct mapping of $v'$ in posed space. From $v_p$, we calculate the 3D Gaussian center $x$ as given in Eq.~\eqref{eq:gaussian_center_from_verts} and 3D rotation $q$ from the 2D rotation as explained in \S \ref{sec:binding_gaussians_to_mesh_surface}.

\subsection{Normal Map from 3D Gaussian}
\label{sec:normal_map_from_3d_gaussian}
Gaussian Splats are generally optimized using RGB photometric loss~\cite{kerbl20233d,qian2023gaussianavatars}. However, we note that this approach results in a poor mesh, low on details. Our goal is to compute details of human surface, \eg, facial attributes, wrinkles and hair [See Supp]. To that end, we take advantage of two crucial facts:
\begin{enumerate}

\item We have explicit representation of vertices($v_t$) and faces ($i_x$) available for each Gaussian to obtain its normal without ambiguity from Eq.~\eqref{eq: normal_from_faces}.

\item SOTA methods like ECON~\cite{xiu2022econ,xiu2021icon,saito2020pifuhd} rely on normal map prediction from RGB to produce SOTA results.
\end{enumerate}

One can therefore use the depth gradient $\nabla\text{Depth}$ in order to compute the surface normals. However, such measurements tend to inherently noisy as it relies on the alpha blending of the gaussians which can introduce noise [see Supp]. On the other hand, the normal image $\bar{I}_{\hat{n}}$ should also equal to the aligned normals obtained from the posed vertices $\{v_p\}$.
We first compute the mesh/Gaussian normals using Eq.~\eqref{eq: normal_from_faces}.

\begin{equation}
    \hat{n} = \frac{(v_p[i_x[1]] - v_p[i_x[0]]) \times (v_p[i_x[2]] - v_p[i_x[0]])}{\|(v_p[i_x[1]] - v_p[i_x[0]]) \times (v_p[i_x[2]] - v_p[i_x[0]])\|}
    \label{eq: normal_from_faces}
\end{equation}
where
\( v_p[i_x[j]] \) refers to the \( j \)-th posed vertex of the triangle in the face \( i_x \). \\

In order to obtain the normal map image from the estimated normals of Eq.~\eqref{eq: normal_from_faces}, we again make use of the Gaussian splatting rasterizer. In order to preserve smoothness and accuracy, the Gaussian Splatting rasterizer already provides a highly efficient approach for normal map computation. For that purpose, we encode the normal $\hat{n}$ into a second spherical harmonics function $SH_{\hat{n}}$ of degree $0$ by representing the components of the normal as: $\hat{n}_r$, $\hat{n}_g$ and $\hat{n}_b$ related to the $rgb$ values of the rasterizer.
The Spherical harmonics again operates on the hash encoding of the vertices $\{v_p\}$ for efficiency and can be evaluated as:
\begin{equation}
    {SH_{\hat{n}}} = \frac{1}{\sqrt{4\pi}} \begin{bmatrix} \hat{n}_r \\ \hat{n}_g \\ \hat{n}_b \end{bmatrix}
\end{equation}

Thus we obtain the final normal prediction $I_{\hat{n}}$ using another pass of the Gaussian rasterizer with $SH_{\hat{n}}$. The final normal loss is therefore obtained as: $\mathcal{L}_{normal} = \bar{I}_{\hat{n}} - I_{\hat{n}}$. We obtain the ground truth normal map directly using a pre-trained pix2pixHD~\cite{wang2018high} network in order to obtain ($\bar{I}_{\hat{n}}$) for every frame $t$. 



\subsection{Training}
Given a set of training images and input poses, we learn our Gaussian Human Template Model iHuman by optimizing the following objective function:
\begin{equation}
    \mathcal{L} = \mathcal{L}_{rgb} + \mathcal{L}_{normal} + \mathcal{L}_{reg}
\end{equation}
where $\mathcal{L}_{rgb}$ is the photometric loss,  $\mathcal{L}_{normal}$ is the normal map loss and $\mathcal{L}_{reg}$ is the 3D regularization term for normal consistency. 
We employ a combination of $\ell_1$ and D-SSIM term Eq.~\eqref{eq:gaussian_loss} for both the $\mathcal{L}_{rgb}$ and the $\mathcal{L}_{normal}$, with the hyperparameter $\lambda = 0.2$:
\begin{equation}
\label{eq:gaussian_loss}
 \mathcal{L} = (1 - \lambda)\mathcal{L}_1 + \lambda\mathcal{L}_{D_{\text{SSIM}}}.
\end{equation}

\section{Experiments}
\subsection{Implementation Details}
We use PyTorch~\cite{Paszke2010pytorch} for the implementation and we choose Adam~\cite{Bengio2015Adam} as the optimizer. We conduct all experiments on a single NVIDIA RTX 4090.
We use standard skinned human body template model, SMPL \cite{matthew2015smpl} as initial mesh template and also use its blend skinning weights. We upsample the mesh to obtain 165K faces in order to initialize our model. To obtain the ground truth normal maps for normal supervision, we use same pix2pixHD~\cite{wang2018high} network as used in PIFuHD \cite{saito2020pifuhd}.
Our method runs at 20 iterations per second (optimization on 20 images in 1 second for 1 epoch) during training with  $>100$ fps during inference.

\subsection{Datasets and Baselines}
\textbf{Datasets.} We conduct experiments on 3 different datasets.
\paragraph{PeopleSnapshot \cite{DBLP:conf/cvpr/AlldieckMXTP18}.}
It comprises of various monocular RGB videos of different subjects recorded in natural settings. In these videos, individuals assume an A-pose and rotate in place facing a stationary camera. We follow the same evaluation protocol as Instant Avatar \cite{jiang2022instantavatar} by training our model with the pose parameters optimized by Anim-NeRF \cite{chen2021animatable}. We keep the poses frozen throughout training for a fair comparison.

\paragraph{UBC-Fashion \cite{zablotskaia2019dwnet}.}
PeopleSnapshot \cite{DBLP:conf/cvpr/AlldieckMXTP18} contains tight clothings and all subjects assume an A pose. In order to evaluate our method on in-the-wild long clothing variations, we use videos from UBC-Fashion. As shown in Fig. \ref{fig:ubc_fashion_view_n_3d}, the subjects in UBC Fashion~\cite{zablotskaia2019dwnet} turn around in-front of a stationary camera in loose clothing. We use the SOTA 3D human pose estimator ReFit \cite{wang2023refit} to obtain the SMPL poses. As the obtained poses can be misaligned, we enable pose optimization during training with our method.

\paragraph{Multi-Garment dataset \cite{bhatnagar2019multi}.}
Due to lack of high-quality geometry data of human body in general clothing, we synthesize several sequences from Multi-Garment Network (MGN)~\cite{bhatnagar2019multi} dataset. The MGN dataset features 3D scanned models of the human body complete with textures, along with corresponding SMPL-D models that are registered for use in animation. For the creation of the videos, we chose 4 human body models of different body types and clothing variations. These models were animated based on motion sequences from the People-Snapshot dataset~\cite{DBLP:conf/cvpr/AlldieckMXTP18}, where subjects rotate in an A-pose.

The synthetic data from the MGN dataset are mainly used to evaluate the quality of the 3D reconstructions. PeopleSnapshot is used for quantitative evaluation of novel view synthesis. Finally we use both PeopleSnapshot and UBC-Fashion for qualitative evaluation of novel view synthesis and 3D reconstructions. 
\newline

\noindent\textbf{Baselines.} We use the recent works 
\textbf{GART} \cite{lei_gart_2023}, \textbf{Anim-NeRF} \cite{chen2021animatable} and \textbf{Instant-Avatar} \cite{jiang2022instantavatar} as our baselines. GART represents the human body in canonical pose represented by the 3D Gaussian parameters and is therefore relatively fast.
\textbf{Anim-NeRF} \cite{chen2021animatable}
utilizes a multi-layer perceptron (MLP) based Neural Radiance Fields (NeRF) \cite{mildenhall2020nerf} to represent human features in a canonical domain and therefore naturally requires longer to optimize. 
Finally, \textbf{Instant-Avatar} \cite{jiang2022instantavatar}
also employs NeRF\cite{mildenhall2020nerf} based method that improves the speed of Anim-NeRF. They achieve this speed up by the use of Instant-NGP\cite{thomas2022instant} for radiance field representation and rendering, Fast SNARF\cite{chen2023fast} for articulation and by the use of occupancy grid for empty space skipping.

\subsection{Evaluations}

\subsubsection{3D Mesh reconstruction.}
SOTA 3D human mesh reconstruction based methods such as vid2Avatar \cite{guo2023vid2avatar}, selfRecon \cite{jiang2022selfrecon} requires more than 1 day of training for a single avatar reconstruction. In contrast, our iHuman is orders of times faster (15 seconds of training time) and uses lower memory. In our experimental setup, we limit the running time of all methods to \textbf{hard limit of 5 minutes}. So, we compare our method with current radiance-field based methods; GART, Anim-NeRF with relatively faster training time and from which meshes can be extracted. Specifically, we use marching cubes \cite{lorensen1998marching} for mesh extraction from Anim-NeRF and poisson reconstruction \cite{kazhdan2006poisson} on the point cloud obtained from GART to obtain the meshes.

\begin{figure}[h]
    \centering
    \includegraphics[width=1\linewidth]{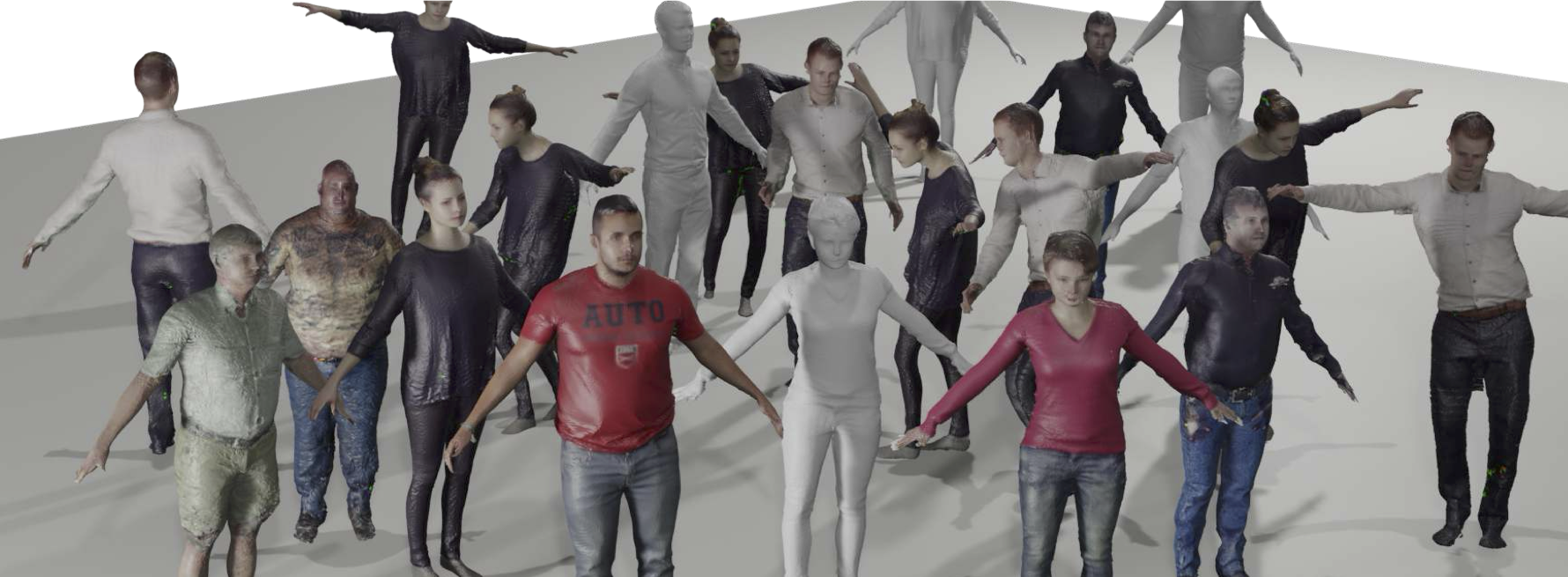}
    \caption{Qualitative results: we obtain fully rigged colored mesh using iHuman since the reconstructed mesh share the same topology with SMPL body model. The obtained meshes are watertight and accurate.}
    \label{fig:crowd_of_3d_human}
\end{figure}
\paragraph{Metrics.} We use bi-directional vertex to vertex (v2v) distances (in mm) calculated by uniformly sampling the predicted and the ground truth mesh on the MGN dataset. Following standard evaluation protocols, our first step involves aligning the centers of the meshes and fixing their scales. Following this alignment process, vertex-to-vertex (v2v) computations are conducted while the meshes are in the canonical T-pose configuration. We also report normal consistency for surface reconstruction comparison. 

\begin{table}[h]
\vspace{-5mm}
\centering
\resizebox{\columnwidth}{!}{%
\begin{tabular}{|l|cccc|cccc|cccc|}
\hline
\textbf{Subject} & \multicolumn{4}{c|}{\textbf{Ours}} & \multicolumn{4}{c|}{\textbf{Anim-NeRF}} & \multicolumn{4}{c|}{\textbf{GART}} \\
 & \textbf{v2v} & \textbf{NC}&\textbf{PSNR} & \textbf{LPIPS} & \textbf{v2v} & \textbf{NC}&\textbf{PSNR} & \textbf{LPIPS} & \textbf{v2v} & \textbf{NC}&\textbf{PSNR} &\textbf{LPIPS}  \\
\hline
Subject-1 & 11.15 & 0.0309& 30.50 & 0.0172 & 71.22 & 0.3294 & 23.16 & 0.0745 & 147.81 & 0.0401 & 35.24 & 0.0140\\
Subject-2 & 12.81 & 0.0327& 29.18 & 0.0267 & 84.62 & 0.3795 & 24.24 & 0.0744 & 138.37 & 0.0457 & 35.72 & 0.0160\\
Subject-3 & 11.78 & 0.0301& 31.63 & 0.0159 & 70.41 & 0.3310 & 24.21 & 0.0691 & 134.43 & 0.0429 & 36.27 & 0.0146\\
Subject-4 & 13.12 & 0.0302& 31.58 & 0.0178 & 68.52 & 0.3428 & 24.04 & 0.0667 & 152.99 & 0.0370 & 35.66 & 0.0162\\
\hline
\end{tabular}%
}
\caption{\textbf{Numerical evaluation on MGN}. We report v2v error (mm), mesh normal consistency (NC), PSNR and LPIPS by our method, Anim-NeRF \cite{chen2021animatable} and GART \cite{lei_gart_2023}. }
\label{tab:3d_error_on_mg}
\vspace{-10mm}
\end{table}


\paragraph{Comparisons.} The quantitative results are shown in Tab.~\ref{tab:3d_error_on_mg}. Our method achieves significantly better results on surface reconstruction compared to GART and Anim-NeRF demonstrating the superiority of our approach in producing accurate geometry.

\begin{figure}[h]
    \centering
    \includegraphics[width=1\linewidth]{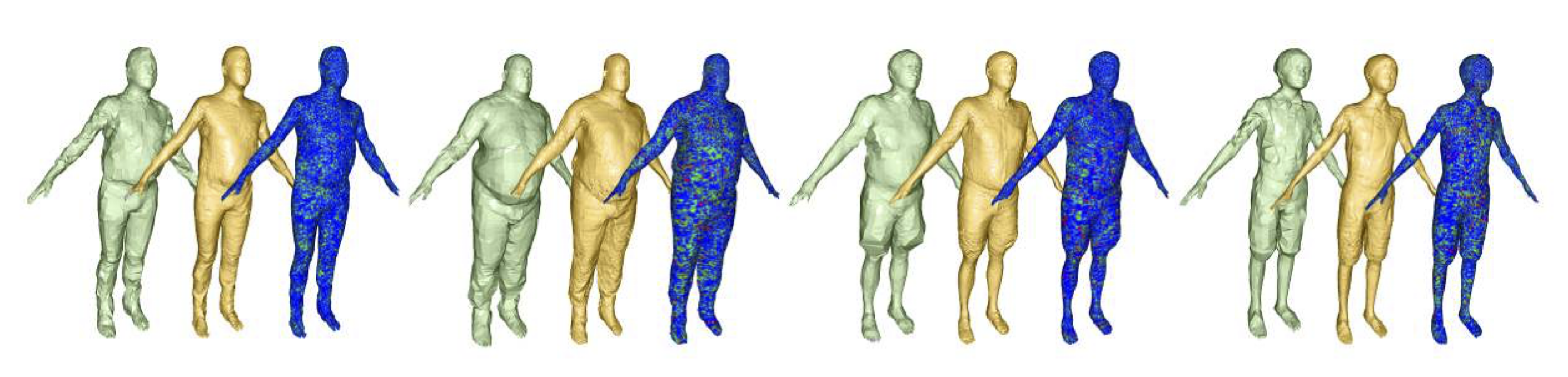}
    \caption{Results on MGN dataset \cite{bhatnagar2019multi}: We compare the ground truth shapes (green) and prediction (yellow) along with corresponding error heatmaps with respect to ground truth shapes (blue represents errors \(\leq 1\) cm and red represents errors \(\geq 3\) cm).}
    \label{fig:v2v_mg}
\end{figure}

\begin{figure}[h]
    \begin{minipage}[t]{0.48\linewidth}
        \centering
        \includegraphics[width=\linewidth]{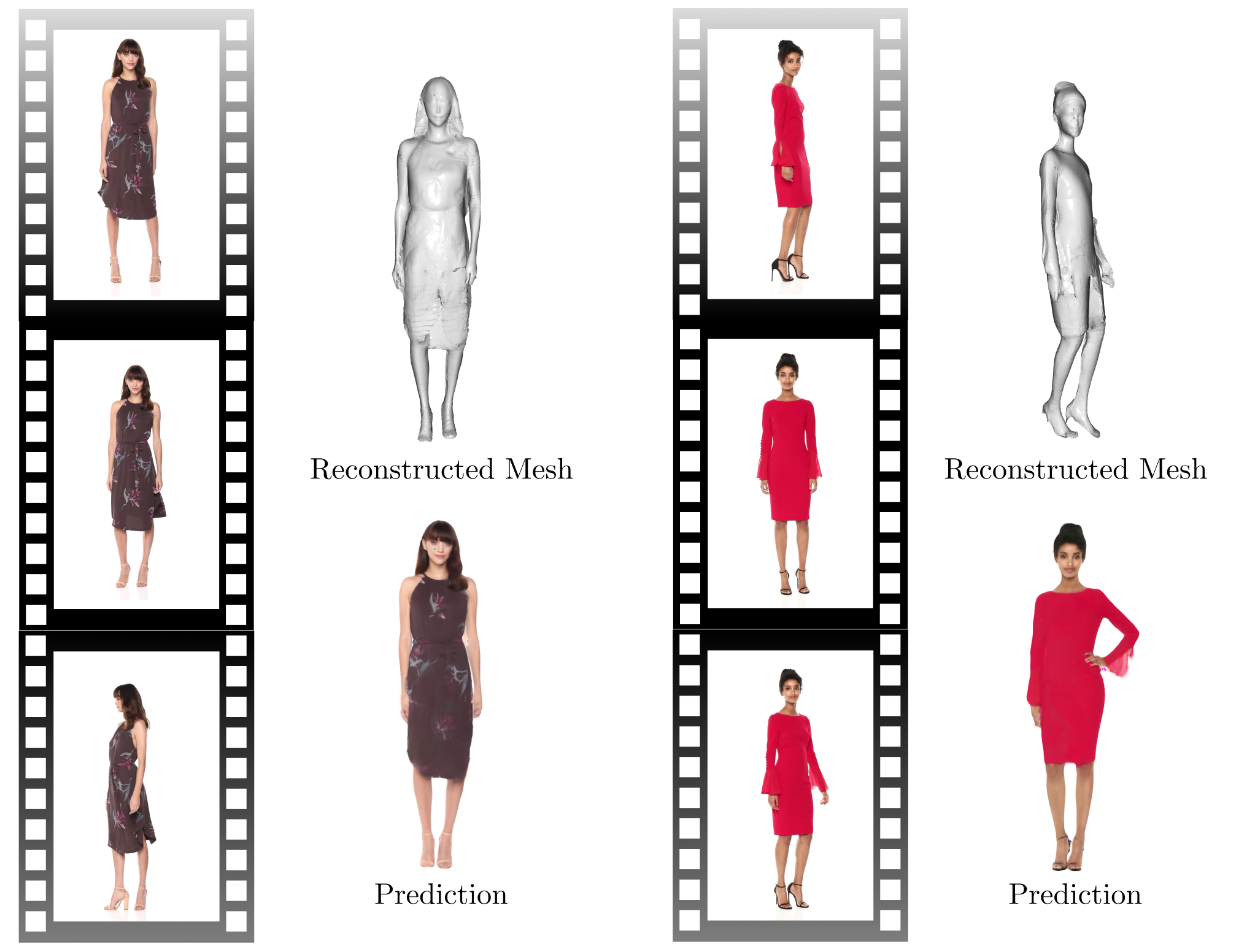}
        \caption{\textbf{View Synthesis and 3D Mesh Reconstruction on challenging UGB-Fashion dataset.} Faithful reconstruction on UBC-Fashion shows robustness of our method on variety of clothing and poses.}
        \label{fig:ubc_fashion_view_n_3d}
    \end{minipage}%
    \hfill
    \begin{minipage}[t]{0.48\linewidth}
        \centering
        \includegraphics[width=\linewidth]{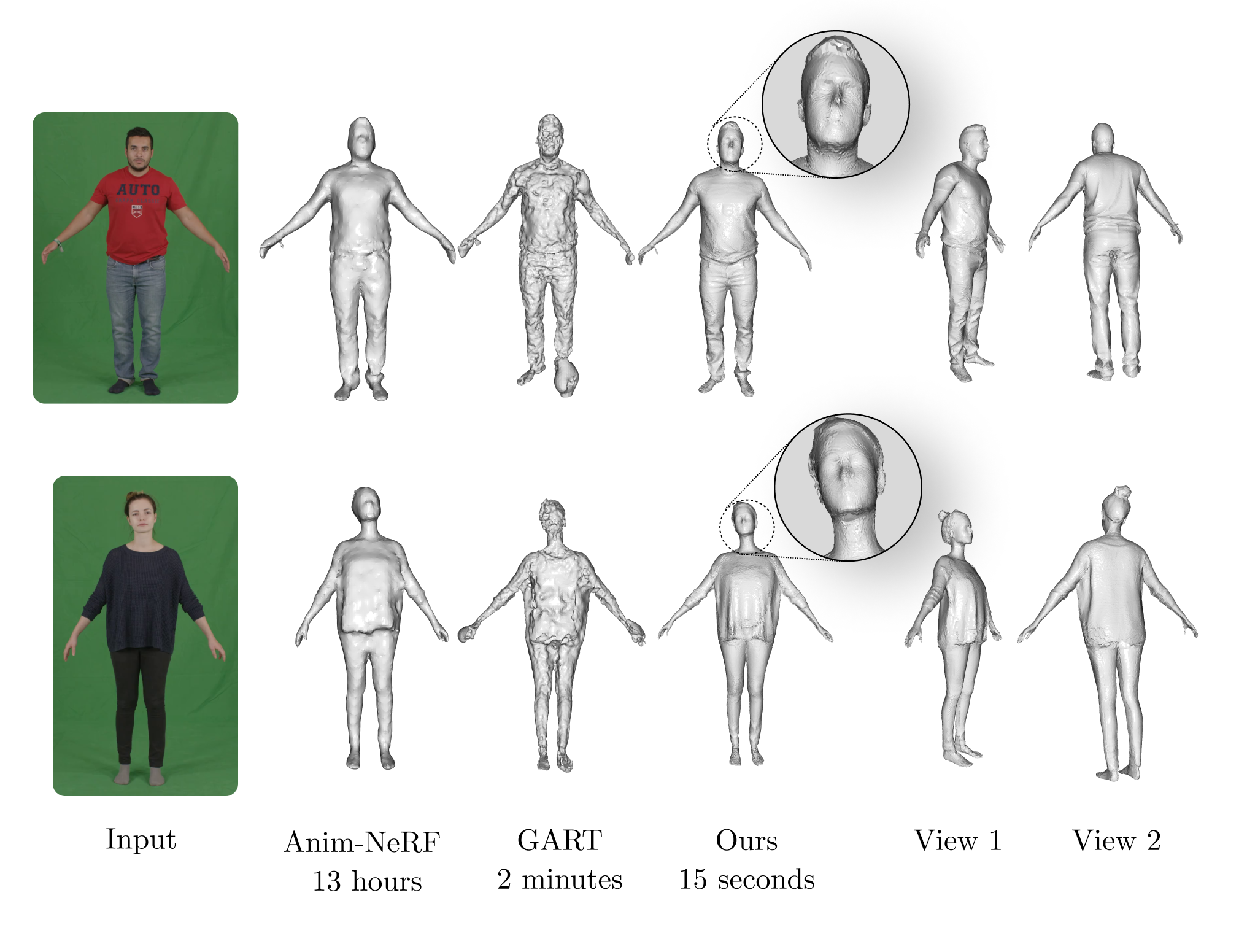}
         \caption{\textbf{Qualitative results of 3D Mesh reconstruction on PeopleSnapshot \cite{DBLP:conf/cvpr/AlldieckMXTP18}.} Our method produces high fidelity mesh even capturing subtle facial details like hair, ear in 15 seconds of computational budget.}
        \label{fig:comparative_3d_on_peoplesnapshot}
    \end{minipage}
\end{figure}

In Fig.~\ref{fig:v2v_mg}, we show side by side comparison of the ground truth subject and the predicted mesh of our method along with the v2v error heatmap. Our iHuman robustly handles mesh reconstruction for different clothing and body types for different subjects.
In Fig.~\ref{fig:crowd_of_3d_human} we show some example reconstructions on PeopleSnapshot dataset and MGN dataset. We show more qualitative results along with comparison in Fig.~\ref{fig:comparative_3d_on_peoplesnapshot}. In only 15s, our method recovers high frequency details such as face structures while other methods struggle with coarse body geometry. For evaluation on more in-the-wild dataset, we show reconstruction on UBC-Fashion dataset in Fig.~\ref{fig:ubc_fashion_view_n_3d}. We provide more evaluations in the supplementary.

\vspace{-1.1cm}

\subsubsection{Novel View Synthesis.}
We quantitatively evaluate our method on novel view synthesis and report PSNR, SSIM, and LPIPS metrics on the test frames of PeopleSnapshot.
We note that all of our baselines require masked input image sequences and the corresponding SMPL pose parameters, which are costly to obtain. Thus, iHuman benefits greatly by requiring less video frames in two ways: i) less amount of pre-processing ii) faster training. In this experiment, we limit the training time budget to maximum of 5 minutes for all the methods. 
\vspace{-0.5mm}
\begin{table}[h]
\centering
\resizebox{\columnwidth}{!}{%
\begin{tabular}{lcccccccccccc}
\hline
\multicolumn{1}{c}{\multirow{2}{*}{Methods}} & \multicolumn{3}{l}{male-3} & \multicolumn{3}{l}{male-4} & \multicolumn{3}{l}{female-3} & \multicolumn{3}{l}{female-4} \\
\multicolumn{1}{c}{}                         & PSNR    & SSIM   & LPIPS   & PSNR    & SSIM    & LPIPS    & PSNR    & SSIM    & LPIPS    & PSNR    & SSIM    & LPIPS    \\
\hline
Anim-Nerf(6 views, 5 mins) &    22.01     &  0.9211     &    0.0810     &   22.60      &    0.9285     &     0.0826     &      22.18   &   0.9370   &   0.0686       &   24.44      &     0.9372    &    0.0524      \\
Instant Avatar(6 views, 5 mins)                    &     15.21    &  0.8228      &  0.2653       &    16.61     &   0.8035     &    0.3010      &    16.30    &  0.8210       &    0.2693      &     21.88    &   0.9221      &     0.0880     \\
GART(6 views, 100 secs)                &    \colorbox{Apricot}{27.28}     &    \colorbox{Apricot}{0.9593}    &  \colorbox{Goldenrod}{0.0393}       &    \colorbox{Apricot}{25.84}     &    \colorbox{Apricot}{0.9527}     &     \colorbox{Goldenrod}{0.0543}     &     \colorbox{Apricot}{23.30}    &   \colorbox{Apricot}{0.9440}      &     \colorbox{Goldenrod}{0.0548}     &     \colorbox{Apricot}{27.18}    &    \colorbox{Apricot}{0.9593}     &      \colorbox{Goldenrod}{0.0339}    \\
Ours(6 views, 6 secs)   &  \colorbox{Goldenrod}{25.27}  & \colorbox{Goldenrod}{0.9483}  & \colorbox{Apricot}{0.0301}   &  \colorbox{Goldenrod}{22.64}  &  \colorbox{Goldenrod}{0.9283}   &  \colorbox{Apricot}{0.0525}  &  \colorbox{Goldenrod}{22.44} & \colorbox{Goldenrod}{0.9368} &  \colorbox{Apricot}{0.0426} &  \colorbox{Goldenrod}{25.08}  &\colorbox{Goldenrod}{0.9471}  &  \colorbox{Apricot}{0.0332}        \\
\hline
Anim-Nerf(12 views, 5 mins)    &  23.81    &   0.9847  &   0.0624   &   23.10 & 0.9333  &  0.0789  &   22.38  &  0.9384  &   0.0623  &  25.87  &     0.9466 & 0.0463    \\
Instant Avatar(12 views, 5 mins)                &     22.24    &   0.9226     &     0.1017    &    17.31    &   0.8152      &      0.2774   &    21.75    &    0.9265     &    0.0783     &   19.19      &   0.8789     &    0.1598      \\
GART(12 views, 105 secs) & \colorbox{Apricot}{29.58}&\colorbox{Apricot}{0.9733} & \colorbox{Goldenrod}{0.0315} & \colorbox{Apricot}{25.79} & \colorbox{Apricot}{0.9540} & \colorbox{Goldenrod}{0.0572} & \colorbox{Apricot}{25.15} & \colorbox{Apricot}{0.9597} & \colorbox{Goldenrod}{0.0429} & \colorbox{Apricot}{28.89} & \colorbox{Apricot}{0.9664} & \colorbox{Goldenrod}{0.0327} \\
Ours(12 views, 12 secs) &  \colorbox{Goldenrod}{26.84}  &  \colorbox{Goldenrod}{0.9586}  & \colorbox{Apricot}{0.0219} &  \colorbox{Goldenrod}{24.93}  & \colorbox{Goldenrod}{0.9455}  & \colorbox{Apricot}{0.0372} & 
\colorbox{Goldenrod}{23.22}  & \colorbox{Goldenrod}{0.9444}  & \colorbox{Apricot}{0.0359} & \colorbox{Goldenrod}{26.01} & \colorbox{Goldenrod}{0.9549} &  \colorbox{Apricot}{0.0281} \\   
\hline
Anim-Nerf(20 views, 5 mins) &    23.46     &   0.9288     &    0.0680     &     23.14    &    0.9340     &   0.0798       &       23.91  &    0.9491     &   0.0568       &     24.92    &   0.9408      &   0.0494       \\
Instant Avatar(20 views, 5 mins)  &    26.68     &   0.9531     &   0.0333      &     24.14    &    0.9383     &    0.0568     &    22.52     &    0.9306     &    0.0784      &    26.25     &     0.9516    &    0.0238      \\
GART(20 views, 110 secs)  &   \colorbox{Apricot}{29.99}      &   \colorbox{Apricot}{0.9760}     &     \colorbox{Goldenrod}{0.0327}   &     \colorbox{Apricot}{27.07}    &    \colorbox{Apricot}{0.9635}     &     \colorbox{Goldenrod}{0.0537}     &   \colorbox{Apricot}{25.60}      &     \colorbox{Apricot}{0.9623}    &   \colorbox{Goldenrod}{ 0.0427}      &    \colorbox{Apricot}{28.78}     &    \colorbox{Apricot}{0.9711}     &    \colorbox{Goldenrod}{0.0321}      \\
Ours(20 views, 20 secs)  & \colorbox{Goldenrod}{27.48}  & \colorbox{Goldenrod}{0.9616} &  \colorbox{Apricot}{0.0196} & \colorbox{Goldenrod}{25.67}  & \colorbox{Goldenrod}{0.9506} & \colorbox{Apricot}{0.0337} & \colorbox{Goldenrod}{23.58} & \colorbox{Goldenrod}{0.9478} & \colorbox{Apricot}{0.0330} &  \colorbox{Goldenrod}{27.20} & \colorbox{Goldenrod}{0.9631} & \colorbox{Apricot}{0.0244} \\   
\hline
\end{tabular}%
}
\caption{\textbf{Qualitative Comparison with SoTA on the PeopleSnapshot\cite{DBLP:conf/cvpr/AlldieckMXTP18} dataset}. We report PSNR, SSIM, and LPIPS between real images and the images generated by Anim-NeRF \cite{chen2021animatable}, InstantAvatar\cite{jiang2022instantavatar} and GART \cite{lei_gart_2023} with computational budget of maximum 5 minutes. We compare all three methods and ours at different number of inputs sequences (views). The \colorbox{Apricot}{best} and \colorbox{Goldenrod}{second best} methods on each metrics are marked on the table.}
\label{tab:quantitative_novel_view}
\vspace{-5mm}
\end{table}

In Tab \ref{tab:quantitative_novel_view}, we show novel view synthesis results for the proposed method and the baselines under different number of views. Our iHuman method achieves better LPIPS compared to all the baselines and report second best PSNR.

In Fig.~\ref{fig:novel_view_20_views}, we show novel view and novel pose synthesis on PeopleSnapshot. The ability of our method to accurately model surface geometry helps in better novel poses without artifacts. For evaluation on more in-the-wild setting, we show results on UBC-Fashion in Fig. \ref{fig:ubc_fashion_view_n_3d}.

\subsubsection{Ablation}
To study the effectiveness of our mesh binding strategy, hash based SH, displacement encoder and our novel normal map prediction on 3D reconstruction and novel view synthesis, we conduct the following ablations:
i) removing the hash based SH Encoder
ii) removing the hash based Displacement Encoder
iii) removing the mesh binding of the Gaussians and
iv) removing the normal map supervision.

\begin{figure}[h]
    \centering
    \begin{minipage}{0.48\linewidth}
        \centering
        \includegraphics[width=\linewidth]{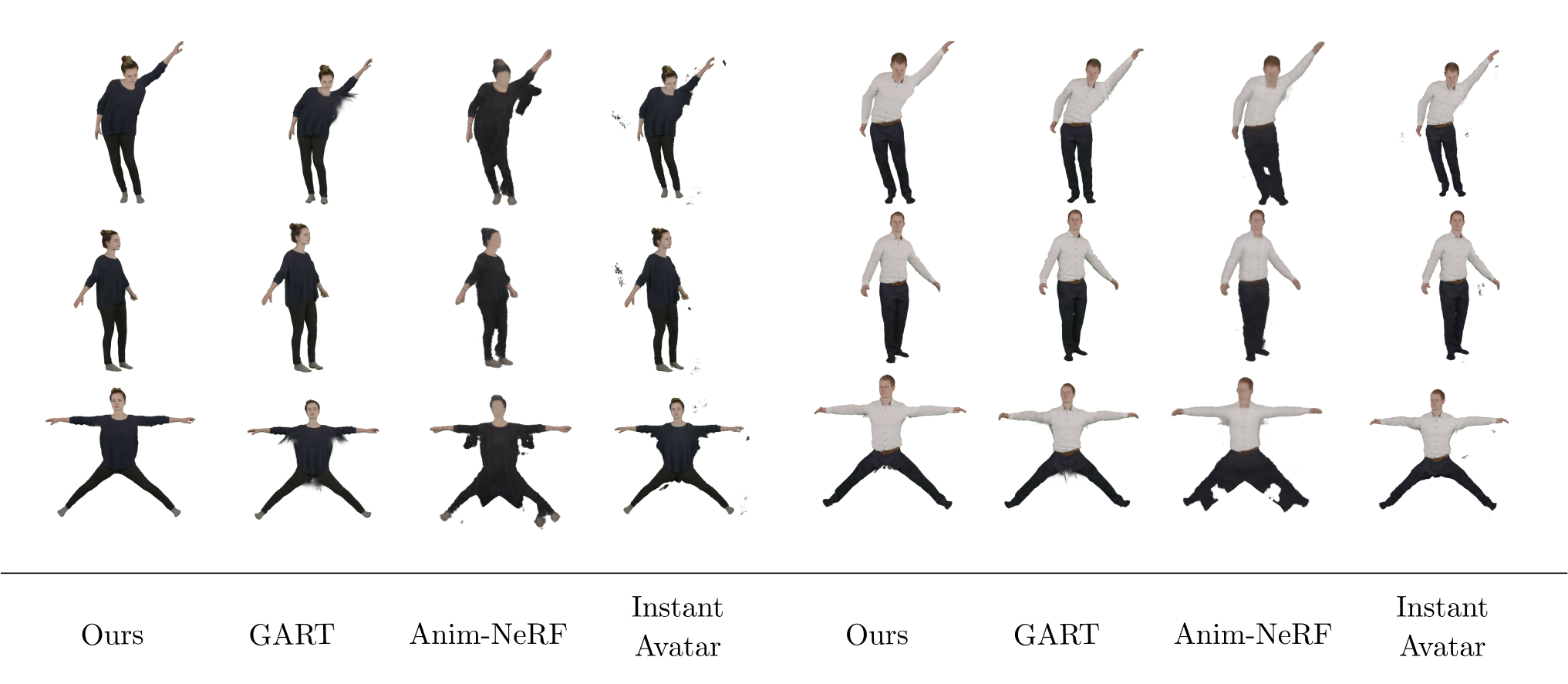}
            \caption{\textbf{ Results on PeopleSnapshot \cite{alldieck2018detailed} for 20 frames.} Our method is robust to poses as it does not contain artifacts even in novel poses. Better viewed zoomed.}
        \label{fig:novel_view_20_views}
    \end{minipage}
    \hfill
    \begin{minipage}{0.48\linewidth}
        \centering
        \includegraphics[width=0.47\linewidth]{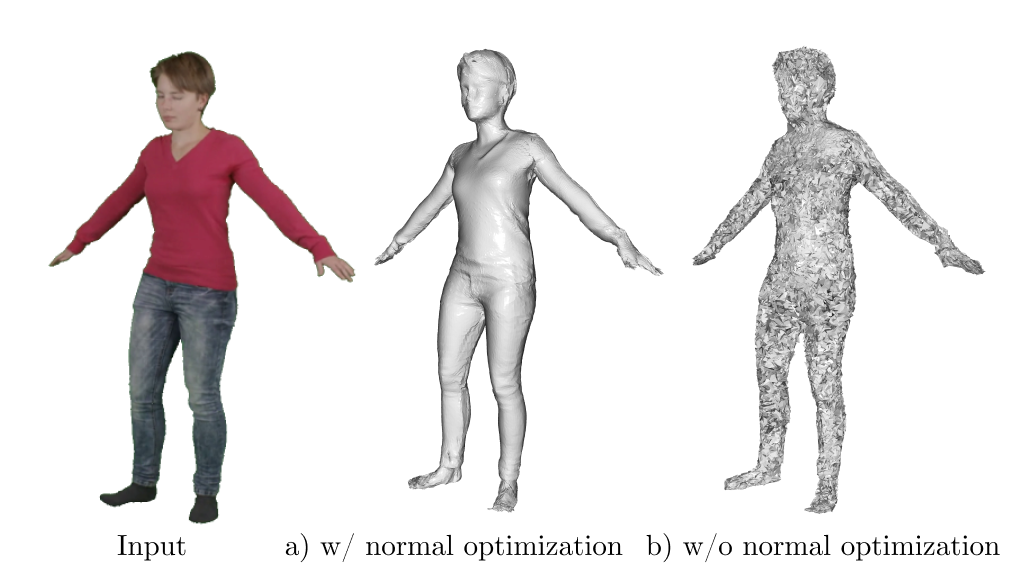}
        \hfill
        \includegraphics[width=0.47\linewidth]{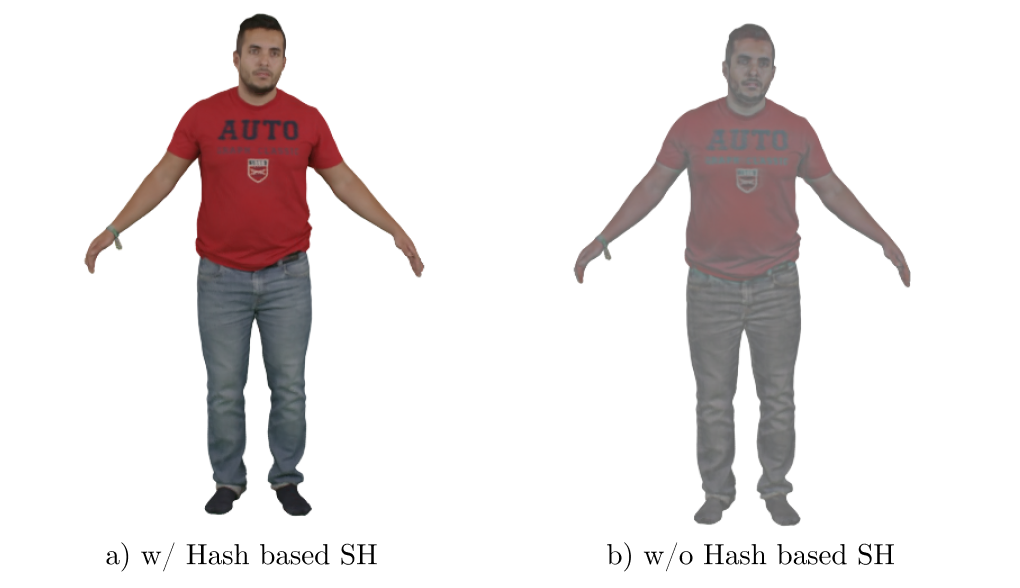}
        \caption{Without normal supervision surface details are not captured (left). Without Hash based SH Encoder, rendered colors are inaccurate (right). Better viewed zoomed.}
        \label{fig:ablation_w_n_wo_normal}
    \end{minipage}
\end{figure}

\begin{table}[h]
    \begin{minipage}[t]{0.48\linewidth} 
      \centering
      \small 
      \caption{\textbf{Ablation Study for Novel View Synthesis.} We evaluate novel view synthesis by disabling key components. The results are averaged on 3 subjects of PeopleSnapshot \cite{alldieck2018detailed}.}
      \label{tab:ablation_novel_view}
      \begin{tabular}{lcc}
      \toprule
      Methods & LPIPS & PSNR \\
      \midrule
      Full & 0.0271 & 26.08 \\
      w/o SH Encoder & 0.0341 & 24.55 \\
      w/o Displacement Encoder & 0.0344 & 25.03 \\
      w/o Mesh Binding & 0.0463 & 24.68 \\
      w/o Normal Loss & 0.0523 & 24.64 \\
      \bottomrule
      \end{tabular}
    \end{minipage}%
    \hfill 
    \begin{minipage}[t]{0.48\linewidth} 
      \centering
      \small 
      \caption{\textbf{Ablation Study for 3D Reconstruction.} We evaluate 3D reconstruction performance by disabling key components of our method. The results are averaged on 4 subjects of MGN~\cite{bhatnagar2019multi}.}
      \label{tab:ablation_3d}
      \begin{tabular}{lcc}
      \toprule
      Methods & v2v & NC \\
      \midrule
      Full & 12.21 & 0.0310 \\
      w/o SH Encoder & 12.82 & 0.0311 \\
      w/o Displacement Encoder & 19.75 & 0.5918 \\
      w/o Mesh Binding & 27.51 & 0.7303 \\
      w/o Normal Loss & 20.82 & 0.5725 \\
      \bottomrule
      \end{tabular}
    \end{minipage}
\end{table}
We show 3D reconstruction results in Tab.~\ref{tab:ablation_3d}.
We observe that displacement encoder helps in better geometric modeling as shown by the normal consistency loss (NC). Without binding of Gaussians to the mesh, both the v2v error and normal consistency degrades.
As shown in Fig.~\ref{fig:ablation_w_n_wo_normal}, only binding Gaussians to the mesh surface cannot produce accurate geometry of surface details with the photometric loss. This highlights the importance of our approach to encode and optimize normals in Gaussian Splatting~\cite{DBLP:journals/tog/KerblKLD23}. In novel view synthesis, we achieve the best PSNR and LPIPS with our full method as shown in Tab.~\ref{tab:ablation_novel_view}. The Hash based SH Encoder improves convergence even for less number of views in Fig.~\ref{fig:ablation_w_n_wo_normal}.

\begin{figure}[h]
    \centering
    \includegraphics[width=0.48\linewidth]{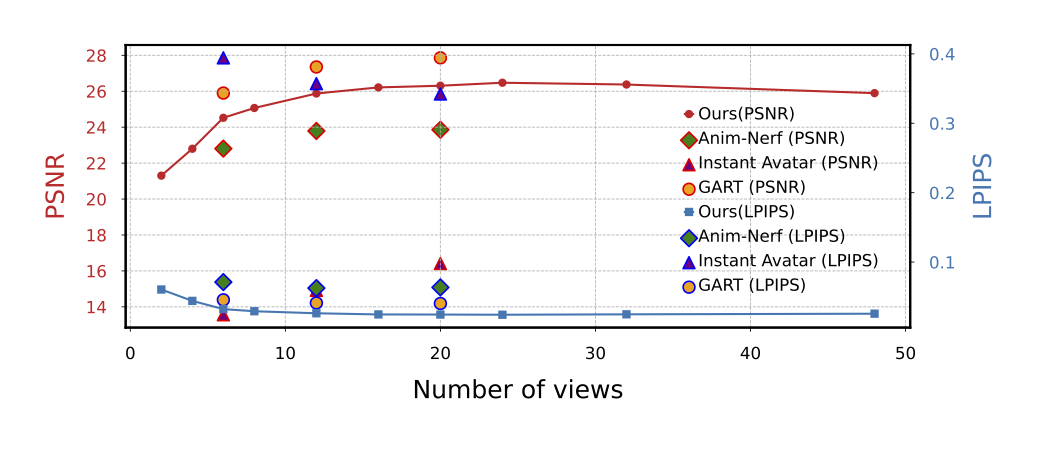}
    \hfill
    \includegraphics[width=0.48\linewidth]{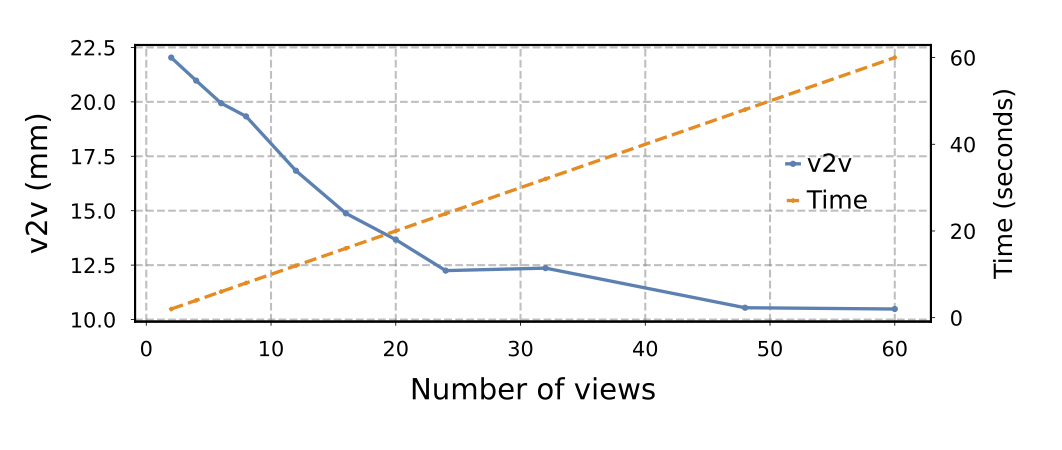}
    \caption{Left: Performance on Novel View Synthesis of iHuman across different input sequences. Right: Performance on 3D Human Mesh Reconstruction of iHuman and training times for varying input views.}
    \label{fig:combined_ablation_studies}
\end{figure}

\textbf{Number of Views}
The performance gain on 3D reconstruction saturates around 50 number of input sequences. For the case of novel view synthesis, 20 views are enough to achieve PSNR of above 25 and very low LPIPS than any other methods under same number of input views in Fig.~\ref{fig:combined_ablation_studies}.

\vspace{-8mm}
\section{Conclusion}
In this work, we proposed a new method to obtain high fidelity animatable human model in record time. We obtain state-of-the-art performance in limited computational budget.
To that end, we used mesh binded Gaussians, explicit normal rasterization and optimization through normal supervision providing fast and accurate results. Through experiments, we also illustrate the need of accurate surface representation, while using Gaussian splats, for faithful rendering under the change in pose.  In the future, we intend to model the temporal smoothness of Gaussian parameters for video frames, that can potentially improve optimization speed and solution even further.




\section*{Acknowledgements}
We would like to express our sincere gratitude to Mr. Kobid Upadhyay for his invaluable assistance in creating the figures for this paper. His contributions have significantly enhanced the clarity and quality of our visual presentations. 

We would also like to thank Alternative Technology (\url{https://alternative.com.np}) for providing us with an RTX 4090 for experimentation, which greatly facilitated our research.

%
%

\bibliographystyle{splncs04}
\bibliography{main}

\begin{thebibliography}{10}
\providecommand{\url}[1]{\texttt{#1}}
\providecommand{\urlprefix}{URL }
\providecommand{\doi}[1]{https://doi.org/#1}

\bibitem{alldieck2018detailed}
Alldieck, T., Magnor, M., Xu, W., Theobalt, C., Pons-Moll, G.: Detailed human avatars from monocular video. In: 2018 International Conference on 3D Vision (3DV). pp. 98--109. IEEE (2018)

\bibitem{DBLP:conf/cvpr/AlldieckMBTP19}
Alldieck, T., Magnor, M.A., Bhatnagar, B.L., Theobalt, C., Pons{-}Moll, G.: Learning to reconstruct people in clothing from a single {RGB} camera. In: {IEEE} Conference on Computer Vision and Pattern Recognition, {CVPR} 2019, Long Beach, CA, USA, June 16-20, 2019. pp. 1175--1186. Computer Vision Foundation / {IEEE} (2019). \doi{10.1109/CVPR.2019.00127}, \url{http://openaccess.thecvf.com/content\_CVPR\_2019/html/Alldieck\_Learning\_to\_Reconstruct\_People\_in\_Clothing\_From\_a\_Single\_RGB\_CVPR\_2019\_paper.html}

\bibitem{DBLP:conf/cvpr/AlldieckMXTP18}
Alldieck, T., Magnor, M.A., Xu, W., Theobalt, C., Pons{-}Moll, G.: Video based reconstruction of 3d people models. In: 2018 {IEEE} Conference on Computer Vision and Pattern Recognition, {CVPR} 2018, Salt Lake City, UT, USA, June 18-22, 2018. pp. 8387--8397. Computer Vision Foundation / {IEEE} Computer Society (2018). \doi{10.1109/CVPR.2018.00875}, \url{http://openaccess.thecvf.com/content\_cvpr\_2018/html/Alldieck\_Video\_Based\_Reconstruction\_CVPR\_2018\_paper.html}

\bibitem{DBLP:conf/eccv/BhatnagarSTP20}
Bhatnagar, B.L., Sminchisescu, C., Theobalt, C., Pons{-}Moll, G.: Combining implicit function learning and parametric models for 3d human reconstruction. In: Vedaldi, A., Bischof, H., Brox, T., Frahm, J. (eds.) Computer Vision - {ECCV} 2020 - 16th European Conference, Glasgow, UK, August 23-28, 2020, Proceedings, Part {II}. Lecture Notes in Computer Science, vol. 12347, pp. 311--329. Springer (2020). \doi{10.1007/978-3-030-58536-5\_19}, \url{https://doi.org/10.1007/978-3-030-58536-5\_19}

\bibitem{DBLP:conf/nips/BhatnagarSTP20}
Bhatnagar, B.L., Sminchisescu, C., Theobalt, C., Pons{-}Moll, G.: Loopreg: Self-supervised learning of implicit surface correspondences, pose and shape for 3d human mesh registration. In: Larochelle, H., Ranzato, M., Hadsell, R., Balcan, M., Lin, H. (eds.) Advances in Neural Information Processing Systems 33: Annual Conference on Neural Information Processing Systems 2020, NeurIPS 2020, December 6-12, 2020, virtual (2020), \url{https://proceedings.neurips.cc/paper/2020/hash/970af30e481057c48f87e101b61e6994-Abstract.html}

\bibitem{bhatnagar2019multi}
Bhatnagar, B.L., Tiwari, G., Theobalt, C., Pons-Moll, G.: Multi-garment net: Learning to dress 3d people from images. In: Proceedings of the IEEE/CVF international conference on computer vision. pp. 5420--5430 (2019)

\bibitem{Bogo:ECCV:2016}
Bogo, F., Kanazawa, A., Lassner, C., Gehler, P., Romero, J., Black, M.J.: Keep it {SMPL}: Automatic estimation of {3D} human pose and shape from a single image. In: Computer Vision -- ECCV 2016. Lecture Notes in Computer Science, Springer International Publishing (Oct 2016)

\bibitem{chen2021animatable}
Chen, J., Zhang, Y., Kang, D., Zhe, X., Bao, L., Jia, X., Lu, H.: Animatable neural radiance fields from monocular rgb videos. arXiv preprint arXiv:2106.13629  (2021)

\bibitem{chen2023fast}
Chen, X., Jiang, T., Song, J., Rietmann, M., Geiger, A., Black, M.J., Hilliges, O.: Fast-snarf: A fast deformer for articulated neural fields. IEEE Transactions on Pattern Analysis and Machine Intelligence  (2023)

\bibitem{chen2021snarf}
Chen, X., Zheng, Y., Black, M.J., Hilliges, O., Geiger, A.: Snarf: Differentiable forward skinning for animating non-rigid neural implicit shapes. In: Proceedings of the IEEE/CVF International Conference on Computer Vision. pp. 11594--11604 (2021)

\bibitem{dragomir2005scape}
Dragomir, A., Praveen, S., Daphne, K., Sebastian, T., Jim, R., James, D.: Scape. ACM Transactions on Graphics ( TOG)  (2005). \doi{10.1145/1073204.1073207}

\bibitem{feng2021learning}
Feng, Y., Feng, H., Black, M.J., Bolkart, T.: Learning an animatable detailed 3d face model from in-the-wild images. ACM Transactions on Graphics (ToG)  \textbf{40}(4),  1--13 (2021)

\bibitem{Gafni_2021_CVPR}
Gafni, G., Thies, J., Zollhofer, M., Niessner, M.: Dynamic neural radiance fields for monocular 4d facial avatar reconstruction. In: Proceedings of the IEEE/CVF Conference on Computer Vision and Pattern Recognition (CVPR). pp. 8649--8658 (June 2021), \url{https://openaccess.thecvf.com/content/CVPR2021/html/Gafni_Dynamic_Neural_Radiance_Fields_for_Monocular_4D_Facial_Avatar_Reconstruction_CVPR_2021_paper.html}

\bibitem{geng2023learning}
Geng, C., Peng, S., Xu, Z., Bao, H., Zhou, X.: Learning neural volumetric representations of dynamic humans in minutes. CVPR  (2023)

\bibitem{goel2023humans}
Goel, S., Pavlakos, G., Rajasegaran, J., Kanazawa, A., Malik, J.: Humans in 4d: Reconstructing and tracking humans with transformers. arXiv preprint arXiv:2305.20091  (2023)

\bibitem{guo2023vid2avatar}
Guo, C., Jiang, T., Chen, X., Song, J., Hilliges, O.: Vid2avatar: 3d avatar reconstruction from videos in the wild via self-supervised scene decomposition. In: Proceedings of the IEEE/CVF Conference on Computer Vision and Pattern Recognition. pp. 12858--12868 (2023)

\bibitem{guédon2023sugar}
Guédon, A., Lepetit, V.: Sugar: Surface-aligned gaussian splatting for efficient 3d mesh reconstruction and high-quality mesh rendering (2023)

\bibitem{He_2021_ICCV}
He, T., Xu, Y., Saito, S., Soatto, S., Tung, T.: Arch++: Animation-ready clothed human reconstruction revisited. In: Proceedings of the IEEE/CVF International Conference on Computer Vision (ICCV). pp. 11046--11056 (October 2021), \url{https://openaccess.thecvf.com/content/ICCV2021/html/He_ARCH_Animation-Ready_Clothed_Human_Reconstruction_Revisited_ICCV_2021_paper.html}

\bibitem{huang2020arch}
Huang, Z., Xu, Y., Lassner, C., Li, H., Tung, T.: Arch: Animatable reconstruction of clothed humans. In: Proceedings of the IEEE/CVF Conference on Computer Vision and Pattern Recognition. pp. 3093--3102 (2020)

\bibitem{jena_splatarmor_2023}
Jena, R., Iyer, G.S., Choudhary, S., Smith, B., Chaudhari, P., Gee, J.: {SplatArmor}: {Articulated} {Gaussian} splatting for animatable humans from monocular {RGB} videos (Nov 2023), \url{http://arxiv.org/abs/2311.10812}, arXiv:2311.10812 [cs]

\bibitem{jiang2022selfrecon}
Jiang, B., Hong, Y., Bao, H., Zhang, J.: Selfrecon: Self reconstruction your digital avatar from monocular video. In: Proceedings of the IEEE/CVF Conference on Computer Vision and Pattern Recognition. pp. 5605--5615 (2022)

\bibitem{jiang2022instantavatar}
Jiang, T., Chen, X., Song, J., Hilliges, O.: Instantavatar: Learning avatars from monocular video in 60 seconds. CVPR  (2023)

\bibitem{DBLP:conf/eccv/JiangYSTR22}
Jiang, W., Yi, K.M., Samei, G., Tuzel, O., Ranjan, A.: Neuman: Neural human radiance field from a single video. In: Avidan, S., Brostow, G.J., Ciss{\'{e}}, M., Farinella, G.M., Hassner, T. (eds.) Computer Vision - {ECCV} 2022 - 17th European Conference, Tel Aviv, Israel, October 23-27, 2022, Proceedings, Part {XXXII}. Lecture Notes in Computer Science, vol. 13692, pp. 402--418. Springer (2022). \doi{10.1007/978-3-031-19824-3\_24}, \url{https://doi.org/10.1007/978-3-031-19824-3\_24}

\bibitem{jiang2023instantnvr}
Jiang, Y., Yao, K., Su, Z., Shen, Z., Luo, H., Xu, L.: Instant-nvr: Instant neural volumetric rendering for human-object interactions from monocular rgbd stream. CVPR  (2023)

\bibitem{kanazawa2017endtoend}
Kanazawa, A., Black, M.J., Jacobs, D., Malik, J.: End-to-end recovery of human shape and pose. IEEE/CVF Conference on Computer Vision and Pattern Recognition  (2017). \doi{10.1109/CVPR.2018.00744}

\bibitem{kazhdan2006poisson}
Kazhdan, M., Bolitho, M., Hoppe, H.: Poisson surface reconstruction. In: Proceedings of the fourth Eurographics symposium on Geometry processing. vol.~7, p.~0 (2006)

\bibitem{kerbl20233d}
Kerbl, B., Kopanas, G., Leimk{\"u}hler, T., Drettakis, G.: 3d gaussian splatting for real-time radiance field rendering. ACM Transactions on Graphics  \textbf{42}(4) (2023)

\bibitem{DBLP:journals/tog/KerblKLD23}
Kerbl, B., Kopanas, G., Leimk{\"{u}}hler, T., Drettakis, G.: 3d gaussian splatting for real-time radiance field rendering. {ACM} Trans. Graph.  \textbf{42}(4),  139:1--139:14 (2023). \doi{10.1145/3592433}, \url{https://doi.org/10.1145/3592433}

\bibitem{Bengio2015Adam}
Kingma, D.P., Ba, J.: Adam: {A} method for stochastic optimization  (2015)

\bibitem{kocabas_hugs_2023}
Kocabas, M., Chang, J.H.R., Gabriel, J., Tuzel, O., Ranjan, A.: {HUGS}: {Human} {Gaussian} {Splats} (Nov 2023), \url{http://arxiv.org/abs/2311.17910}, arXiv:2311.17910 [cs]

\bibitem{kolotouros2019convolutional}
Kolotouros, N., Pavlakos, G., Daniilidis, K.: Convolutional mesh regression for single-image human shape reconstruction. CVPR  (2019)

\bibitem{kwon2021neural}
Kwon, Y., Kim, D., Ceylan, D., Fuchs, H.: Neural human performer: Learning generalizable radiance fields for human performance rendering. Advances in Neural Information Processing Systems  \textbf{34},  24741--24752 (2021)

\bibitem{lei_gart_2023}
Lei, J., Wang, Y., Pavlakos, G., Liu, L., Daniilidis, K.: {GART}: {Gaussian} {Articulated} {Template} {Models} (Nov 2023), \url{http://arxiv.org/abs/2311.16099}, arXiv:2311.16099 [cs]

\bibitem{lewis2023pose}
Lewis, J.P., Cordner, M., Fong, N.: Pose space deformation: a unified approach to shape interpolation and skeleton-driven deformation. In: Seminal Graphics Papers: Pushing the Boundaries, Volume 2, pp. 811--818 (2023)

\bibitem{li2024gaussianbody}
Li, M., Yao, S., Xie, Z., Chen, K., Jiang, Y.G.: Gaussianbody: Clothed human reconstruction via 3d gaussian splatting. arXiv preprint arXiv:2401.09720  (2024)

\bibitem{li_human101_2023}
Li, M., Tao, J., Yang, Z., Yang, Y.: Human101: {Training} 100+{FPS} {Human} {Gaussians} in 100s from 1 {View} (Dec 2023), \url{http://arxiv.org/abs/2312.15258}, arXiv:2312.15258 [cs]

\bibitem{DBLP:conf/eccv/LiTVZGKL22}
Li, R., Tanke, J., Vo, M., Zollh{\"{o}}fer, M., Gall, J., Kanazawa, A., Lassner, C.: {TAVA:} template-free animatable volumetric actors. In: Avidan, S., Brostow, G.J., Ciss{\'{e}}, M., Farinella, G.M., Hassner, T. (eds.) Computer Vision - {ECCV} 2022 - 17th European Conference, Tel Aviv, Israel, October 23-27, 2022, Proceedings, Part {XXXII}. Lecture Notes in Computer Science, vol. 13692, pp. 419--436. Springer (2022). \doi{10.1007/978-3-031-19824-3\_25}, \url{https://doi.org/10.1007/978-3-031-19824-3\_25}

\bibitem{li_animatable_2023}
Li, Z., Zheng, Z., Wang, L., Liu, Y.: Animatable {Gaussians}: {Learning} {Pose}-dependent {Gaussian} {Maps} for {High}-fidelity {Human} {Avatar} {Modeling} (Nov 2023), \url{http://arxiv.org/abs/2311.16096}, arXiv:2311.16096 [cs]

\bibitem{lin2021mesh}
Lin, K., Wang, L., Liu, Z.: Mesh graphormer. ICCV  (2021)

\bibitem{liu2021neural}
Liu, L., Habermann, M., Rudnev, V., Sarkar, K., Gu, J., Theobalt, C.: Neural actor: Neural free-view synthesis of human actors with pose control. ACM transactions on graphics (TOG)  \textbf{40}(6),  1--16 (2021)

\bibitem{lorensen1998marching}
Lorensen, W.E., Cline, H.E.: Marching cubes: A high resolution 3d surface construction algorithm. In: Seminal graphics: pioneering efforts that shaped the field, pp. 347--353 (1998)

\bibitem{marc2023hd}
Marc, H., Lingjie, L., Weipeng, X., Gerard, P.M., Michael, Z., Christian, T.: Hd humans. Proceedings of the ACM on Computer Graphics and Interactive Techniques  (2023). \doi{10.1145/3606927}, \url{https://dl.acm.org/doi/10.1145/3606927}

\bibitem{matthew2015smpl}
Matthew, L., Naureen, M., Javier, R., Gerard, P.M., J., B.M.: Smpl. ACM Transactions on Graphics ( TOG)  (2015). \doi{10.1145/2816795.2818013}, \url{https://dl.acm.org/doi/10.1145/2816795.2818013}

\bibitem{mescheder2018occupancy}
Mescheder, L., Oechsle, M., Niemeyer, M., Nowozin, S., Geiger, A.: Occupancy networks: Learning 3d reconstruction in function space. CVPR  (2019)

\bibitem{mildenhall2020nerf}
Mildenhall, B., Srinivasan, P.P., Tancik, M., Barron, J., Ramamoorthi, R., Ng, R.: Nerf: Representing scenes as neural radiance fields for view synthesis. European Conference on Computer Vision  (2020). \doi{10.1007/978-3-030-58452-8_24}

\bibitem{moon2020i2lmeshnet}
Moon, G., Lee, K.M.: I2l-meshnet: Image-to-lixel prediction network for accurate 3d human pose and mesh estimation from a single rgb image. European Conference on Computer Vision  (2020). \doi{10.1007/978-3-030-58571-6_44}

\bibitem{moreau_human_2023}
Moreau, A., Song, J., Dhamo, H., Shaw, R., Zhou, Y., Pérez-Pellitero, E.: Human {Gaussian} {Splatting}: {Real}-time {Rendering} of {Animatable} {Avatars} (Nov 2023), \url{http://arxiv.org/abs/2311.17113}, arXiv:2311.17113 [cs]

\bibitem{Noguchi_2021_ICCV}
Noguchi, A., Sun, X., Lin, S., Harada, T.: Neural articulated radiance field. In: Proceedings of the IEEE/CVF International Conference on Computer Vision (ICCV). pp. 5762--5772 (October 2021), \url{https://openaccess.thecvf.com/content/ICCV2021/html/Noguchi_Neural_Articulated_Radiance_Field_ICCV_2021_paper.html}

\bibitem{omran2018neural}
Omran, M., Lassner, C., Pons-Moll, G., Gehler, P., Schiele, B.: Neural body fitting: Unifying deep learning and model-based human pose and shape estimation. International Conference on 3D Vision  (2018). \doi{10.1109/3DV.2018.00062}

\bibitem{DBLP:conf/eccv/OsmanBB20}
Osman, A.A.A., Bolkart, T., Black, M.J.: {STAR:} sparse trained articulated human body regressor. In: Vedaldi, A., Bischof, H., Brox, T., Frahm, J. (eds.) Computer Vision - {ECCV} 2020 - 16th European Conference, Glasgow, UK, August 23-28, 2020, Proceedings, Part {VI}. Lecture Notes in Computer Science, vol. 12351, pp. 598--613. Springer (2020). \doi{10.1007/978-3-030-58539-6\_36}, \url{https://doi.org/10.1007/978-3-030-58539-6\_36}

\bibitem{pang_ash_2023}
Pang, H., Zhu, H., Kortylewski, A., Theobalt, C., Habermann, M.: {ASH}: {Animatable} {Gaussian} {Splats} for {Efficient} and {Photoreal} {Human} {Rendering} (Dec 2023), \url{http://arxiv.org/abs/2312.05941}, arXiv:2312.05941 [cs]

\bibitem{DBLP:conf/cvpr/ParkFSNL19}
Park, J.J., Florence, P.R., Straub, J., Newcombe, R.A., Lovegrove, S.: Deepsdf: Learning continuous signed distance functions for shape representation. In: {IEEE} Conference on Computer Vision and Pattern Recognition, {CVPR} 2019, Long Beach, CA, USA, June 16-20, 2019. pp. 165--174. Computer Vision Foundation / {IEEE} (2019). \doi{10.1109/CVPR.2019.00025}, \url{http://openaccess.thecvf.com/content\_CVPR\_2019/html/Park\_DeepSDF\_Learning\_Continuous\_Signed\_Distance\_Functions\_for\_Shape\_Representation\_CVPR\_2019\_paper.html}

\bibitem{Paszke2010pytorch}
Paszke, A., Gross, S., Massa, F., Lerer, A., Bradbury, J., Chanan, G., Killeen, T., Lin, Z., Gimelshein, N., Antiga, L., Desmaison, A., Kopf, A., Yang, E., DeVito, Z., Raison, M., Tejani, A., Chilamkurthy, S., Steiner, B., Fang, L., Bai, J., Chintala, S.: Pytorch: An imperative style, high-performance deep learning library  (2019)

\bibitem{pavlakos2018learning}
Pavlakos, G., Zhu, L., Zhou, X., Daniilidis, K.: Learning to estimate 3d human pose and shape from a single color image. IEEE/CVF Conference on Computer Vision and Pattern Recognition  (2018). \doi{10.1109/CVPR.2018.00055}

\bibitem{Pavlakos_2019_CVPR}
Pavlakos, G., Choutas, V., Ghorbani, N., Bolkart, T., Osman, A.A.A., Tzionas, D., Black, M.J.: Expressive body capture: 3d hands, face, and body from a single image. In: Proceedings of the IEEE/CVF Conference on Computer Vision and Pattern Recognition (CVPR) (June 2019)

\bibitem{peng2021animatable}
Peng, S., Dong, J., Wang, Q., Zhang, S., Shuai, Q., Zhou, X., Bao, H.: Animatable neural radiance fields for modeling dynamic human bodies. In: Proceedings of the IEEE/CVF International Conference on Computer Vision. pp. 14314--14323 (2021)

\bibitem{9710330}
Peng, S., Dong, J., Wang, Q., Zhang, S., Shuai, Q., Zhou, X., Bao, H.: Animatable neural radiance fields for modeling dynamic human bodies. 2021 IEEE/CVF International Conference on Computer Vision (ICCV) pp. 14294--14303 (2021). \doi{10.1109/ICCV48922.2021.01405}

\bibitem{DBLP:conf/cvpr/PengZXWSBZ21}
Peng, S., Zhang, Y., Xu, Y., Wang, Q., Shuai, Q., Bao, H., Zhou, X.: Neural body: Implicit neural representations with structured latent codes for novel view synthesis of dynamic humans. In: {IEEE} Conference on Computer Vision and Pattern Recognition, {CVPR} 2021, virtual, June 19-25, 2021. pp. 9054--9063. Computer Vision Foundation / {IEEE} (2021). \doi{10.1109/CVPR46437.2021.00894}, \url{https://openaccess.thecvf.com/content/CVPR2021/html/Peng\_Neural\_Body\_Implicit\_Neural\_Representations\_With\_Structured\_Latent\_Codes\_for\_CVPR\_2021\_paper.html}

\bibitem{qian2023gaussianavatars}
Qian, S., Kirschstein, T., Schoneveld, L., Davoli, D., Giebenhain, S., Nießner, M.: Gaussianavatars: Photorealistic head avatars with rigged 3d gaussians. arXiv preprint arXiv: 2312.02069  (2023)

\bibitem{qian_3dgs-avatar_2023}
Qian, Z., Wang, S., Mihajlovic, M., Geiger, A., Tang, S.: {3DGS}-{Avatar}: {Animatable} {Avatars} via {Deformable} {3D} {Gaussian} {Splatting} (Dec 2023), \url{http://arxiv.org/abs/2312.09228}, arXiv:2312.09228 [cs]

\bibitem{saito2020pifuhd}
Saito, S., Simon, T., Saragih, J., Joo, H.: Pifuhd: Multi-level pixel-aligned implicit function for high-resolution 3d human digitization. CVPR  (2020)

\bibitem{schonberger2016structure}
Schonberger, J.L., Frahm, J.M.: Structure-from-motion revisited. In: Proceedings of the IEEE conference on computer vision and pattern recognition. pp. 4104--4113 (2016)

\bibitem{shahbazi2023nerf}
Shahbazi, M., Ntavelis, E., Tonioni, A., Collins, E., Paudel, D.P., Danelljan, M., Van~Gool, L.: Nerf-gan distillation for efficient 3d-aware generation with convolutions. arXiv preprint arXiv:2303.12865  (2023)

\bibitem{su2022danbo}
Su, S.Y., Bagautdinov, T.M., Rhodin, H.: Danbo: Disentangled articulated neural body representations via graph neural networks. European Conference on Computer Vision  (2022). \doi{10.48550/arXiv.2205.01666}

\bibitem{su2021anerf}
Su, S.Y., Yu, F., Zollhoefer, M., Rhodin, H.: A-nerf: Articulated neural radiance fields for learning human shape, appearance, and pose. NEURIPS  (2021)

\bibitem{thomas2022instant}
Thomas, M., Alex, E., Christoph, S., Alexander, K.: Instant neural graphics primitives with a multiresolution hash encoding. ACM Transactions on Graphics ( TOG)  (2022). \doi{10.1145/3528223.3530127}, \url{https://dl.acm.org/doi/10.1145/3528223.3530127}

\bibitem{waczynska2024games}
Waczy{\'n}ska, J., Borycki, P., Tadeja, S., Tabor, J., Spurek, P.: Games: Mesh-based adapting and modification of gaussian splatting. arXiv preprint arXiv:2402.01459  (2024)

\bibitem{wang2021neus}
Wang, P., Liu, L., Liu, Y., Theobalt, C., Komura, T., Wang, W.: Neus: Learning neural implicit surfaces by volume rendering for multi-view reconstruction. NEURIPS  (2021)

\bibitem{DBLP:conf/eccv/WangSGT22}
Wang, S., Schwarz, K., Geiger, A., Tang, S.: {ARAH:} animatable volume rendering of articulated human sdfs. In: Avidan, S., Brostow, G.J., Ciss{\'{e}}, M., Farinella, G.M., Hassner, T. (eds.) Computer Vision - {ECCV} 2022 - 17th European Conference, Tel Aviv, Israel, October 23-27, 2022, Proceedings, Part {XXXII}. Lecture Notes in Computer Science, vol. 13692, pp. 1--19. Springer (2022). \doi{10.1007/978-3-031-19824-3\_1}, \url{https://doi.org/10.1007/978-3-031-19824-3\_1}

\bibitem{wang2018high}
Wang, T.C., Liu, M.Y., Zhu, J.Y., Tao, A., Kautz, J., Catanzaro, B.: High-resolution image synthesis and semantic manipulation with conditional gans. In: Proceedings of the IEEE conference on computer vision and pattern recognition. pp. 8798--8807 (2018)

\bibitem{wang2022neural}
Wang, Y., Gao, Q., Liu, L., Liu, L., Theobalt, C., Chen, B.: Neural novel actor: Learning a generalized animatable neural representation for human actors. IEEE Transactions on Visualization and Computer Graphics  (2022). \doi{10.48550/arXiv.2208.11905}, \url{https://arxiv.org/abs/2208.11905v2}

\bibitem{wang2023refit}
Wang, Y., Daniilidis, K.: Refit: Recurrent fitting network for 3d human recovery. In: Proceedings of the IEEE/CVF International Conference on Computer Vision. pp. 14644--14654 (2023)

\bibitem{9879404}
Weng, C., Curless, B., Srinivasan, P.P., Barron, J.T., Kemelmacher-Shlizerman, I.: Humannerf: Free-viewpoint rendering of moving people from monocular video. 2022 IEEE/CVF Conference on Computer Vision and Pattern Recognition (CVPR) pp. 16189--16199 (2022). \doi{10.1109/CVPR52688.2022.01573}

\bibitem{xiu2022econ}
Xiu, Y., Yang, J., Cao, X., Tzionas, D., Black, M.J.: Econ: Explicit clothed humans optimized via normal integration. Computer Vision and Pattern Recognition  (2022). \doi{10.1109/CVPR52729.2023.00057}

\bibitem{xiu2021icon}
Xiu, Y., Yang, J., Tzionas, D., Black, M.J.: Icon: Implicit clothed humans obtained from normals. CVPR  (2022)

\bibitem{xu2021hnerf}
Xu, H., Alldieck, T., Sminchisescu, C.: H-nerf: Neural radiance fields for rendering and temporal reconstruction of humans in motion. NEURIPS  (2021)

\bibitem{yariv2021volume}
Yariv, L., Gu, J., Kasten, Y., Lipman, Y.: Volume rendering of neural implicit surfaces. Advances in Neural Information Processing Systems  \textbf{34},  4805--4815 (2021)

\bibitem{NEURIPS2021_25e2a30f}
Yariv, L., Gu, J., Kasten, Y., Lipman, Y.: Volume rendering of neural implicit surfaces. In: Ranzato, M., Beygelzimer, A., Dauphin, Y., Liang, P., Vaughan, J.W. (eds.) Advances in Neural Information Processing Systems. vol.~34, pp. 4805--4815. Curran Associates, Inc. (2021), \url{https://proceedings.neurips.cc/paper_files/paper/2021/file/25e2a30f44898b9f3e978b1786dcd85c-Paper.pdf}

\bibitem{yu2023monohuman}
Yu, Z., Cheng, W., Liu, X., Wu, W., Lin, K.Y.: Monohuman: Animatable human neural field from monocular video. CVPR  (2023)

\bibitem{yuan2023gavatar}
Yuan, Y., Li, X., Huang, Y., De~Mello, S., Nagano, K., Kautz, J., Iqbal, U.: Gavatar: Animatable 3d gaussian avatars with implicit mesh learning. arXiv preprint arXiv:2312.11461  (2023)

\bibitem{zablotskaia2019dwnet}
Zablotskaia, P., Siarohin, A., Zhao, B., Sigal, L.: Dwnet: Dense warp-based network for pose-guided human video generation. arXiv preprint arXiv:1910.09139  (2019)

\bibitem{zhang2021editable}
Zhang, J., Liu, X., Ye, X., Zhao, F., Zhang, Y., Wu, M., Zhang, Y., Xu, L., Yu, J.: Editable free-viewpoint video using a layered neural representation. ACM Transactions on Graphics (TOG)  \textbf{40}(4),  1--18 (2021)

\bibitem{DBLP:journals/tog/ZhaoJYZWDZZWXY22}
Zhao, F., Jiang, Y., Yao, K., Zhang, J., Wang, L., Dai, H., Zhong, Y., Zhang, Y., Wu, M., Xu, L., Yu, J.: Human performance modeling and rendering via neural animated mesh. {ACM} Trans. Graph.  \textbf{41}(6),  235:1--235:17 (2022). \doi{10.1145/3550454.3555451}, \url{https://doi.org/10.1145/3550454.3555451}

\bibitem{Zheng_2022_CVPR}
Zheng, Z., Huang, H., Yu, T., Zhang, H., Guo, Y., Liu, Y.: Structured local radiance fields for human avatar modeling. In: Proceedings of the IEEE/CVF Conference on Computer Vision and Pattern Recognition (CVPR). pp. 15893--15903 (June 2022), \url{https://openaccess.thecvf.com/content/CVPR2022/html/Zheng_Structured_Local_Radiance_Fields_for_Human_Avatar_Modeling_CVPR_2022_paper.html}

\bibitem{zhu2023trihuman}
Zhu, H., Zhan, F., Theobalt, C., Habermann, M.: Trihuman: A real-time and controllable tri-plane representation for detailed human geometry and appearance synthesis. arXiv preprint arXiv:2312.05161  (2023)

\bibitem{zielonka_drivable_2023}
Zielonka, W., Bagautdinov, T., Saito, S., Zollhöfer, M., Thies, J., Romero, J.: Drivable {3D} {Gaussian} {Avatars} (Nov 2023), \url{http://arxiv.org/abs/2311.08581}, arXiv:2311.08581 [cs]

\bibitem{DBLP:conf/cvpr/ZielonkaBT23}
Zielonka, W., Bolkart, T., Thies, J.: Instant volumetric head avatars. In: {IEEE/CVF} Conference on Computer Vision and Pattern Recognition, {CVPR} 2023, Vancouver, BC, Canada, June 17-24, 2023. pp. 4574--4584. IEEE (2023). \doi{10.1109/CVPR52729.2023.00444}, \url{https://doi.org/10.1109/CVPR52729.2023.00444}

\end{thebibliography}
\end{document}


\title{Supplementary Material}

\titlerunning{iHuman: Instant Digital Humans}

In this supplementary material, we begin by presenting the implementation details of our method by first explaining about the use of Multi-Resolution Hash Encoder as Displacement Encoder and Spherical Harmonics Encoder in detail in \S \ref{sec:hash_encoding}. We then provide more mathematical details on how we bind 3D Gaussians to mesh surface in \S \ref{sec:binding_gaussain_to_mesh_surface}. Our stage wise training strategies are explained in \S \ref{sec:training}.

We present more qualitative restuls and comparision on PeopleSnapshot \cite{alldieck2018detailed} and in-the-wild UBC-Fashion \cite{zablotskaia2019dwnet} dataset in \S \ref{sec:additional_qualitative_results}. Then, we show challenging cases in \S \ref{sec:failure_cases}.

\section{Implementation Details}
\label{sec:implementation_details}
\subsection{Hash Encoding}
\label{sec:hash_encoding}

A single scene reconstructed using 3D gaussian splatting contains a very large number of gaussians. Backpropagating the gradients to every gaussians in every iteration results in poor performance due to which we need an efficient sampling strategy is needed to select only a few gaussians to be updated every iteration. Inspired by \cite{liu2023animatable} we encode the color and displacement for each gaussian using a multi-resolution hash encoder.\\
The color and displacement encoder include a hash encoder followed by a fully-fused MLPs implemented using the tiny-cuda-nn framework \cite{tiny-cuda-nn}.
The implemented hash encoder has 16 levels, hash table size of ${2^{17}}$ with 4 features per entry in the table with a base resolution of 4 and resolution growth factor of 1.5. The fully fused MLPs consist of 2 hidden layer with 64 neurons per layer and uses relu as the activation function. Both the color and displacement encoder take the gaussian position as input and produce a color and displacement value of the same dimension as the input.


\subsection{Binding Gaussian To Mesh Surface}
\label{sec:binding_gaussain_to_mesh_surface}
As we explained in Section 3.3, we bind Gaussians at the centroid of the triangular face. Given face $i_x$, the centroid is given by:

\begin{equation}
\label{eq:gaussian_center_from_verts}
    x=\frac{{i_x[1] + i_x[2] + i_x[3]}}{3}.
\end{equation}

Through this equation, we always maintain the position of Gaussian at the centroid of the face $i_x$. And, we can directly optimize the vertices $V_c$ of canonical mesh $\mathcal{M}=(V_c, F)$.

Similar to SuGaR \cite{guédon2023sugar}, we parameterize the 3D rotation of the Gaussians with only 2 parameters by encoding the rotation in complex 2D rotation form with ($x+iy$). We limit their rotation to local 2D triangular face plane.
We now explain how we convert the local 2D complex rotation to 3D rotation required for Gaussian Rasterizer.

Consider a set of face normals $\mathbf{n}_i$ for each face $i$ of the mesh. We first normalize these normals to obtain a primary direction vector $\mathbf{R}_0$:
\begin{equation}
\mathbf{R}_0 = \frac{\mathbf{n}_i}{\|\mathbf{n}_i\|}.
\end{equation}

The second direction vector $\mathbf{R}_1$ is calculated using the difference between the first two vertices of the triangle, yielding the edge vector $\mathbf{e}_{12}$, which is then normalized:
\begin{equation}
\mathbf{R}_1 = \frac{\mathbf{e}_{12}}{\|\mathbf{e}_{12}\|}.
\end{equation}

The third direction vector $\mathbf{R}_2$ is produced by the cross product of $\mathbf{R}_0$ and $\mathbf{R}_1$, ensuring orthogonality:
\begin{equation}
\mathbf{R}_2 = \frac{\mathbf{R}_0 \times \mathbf{R}_1}{\|\mathbf{R}_0 \times \mathbf{R}_1\|}.
\end{equation}

To incorporate the parameterized 2D rotation defined by complex number of 2 parameters for each Gaussians, we apply them to the base vectors $\mathbf{R}_1$ and $\mathbf{R}_2$ to obtain the rotated axes.

Finally, the rotation matrix $\mathbf{R}$ for each Gaussian distribution is constructed by combining the normalized primary, secondary, and tertiary direction vectors:
\begin{equation}
\mathbf{R} = [\mathbf{R}_0, \mathbf{R}_1, \mathbf{R}_2].
\end{equation}

These rotation matrices $\mathbf{R}$ are used to orient the Gaussians so that they follow the orientation of the mesh surface.

As scaling parameter $S$, we use:
\begin{equation}
    S=(s_1, s_2, s_3)
\end{equation}
where, $s_1=\epsilon$, $s_2$ and $s_3$ are learnable parameters. $s_1$ corresponds with the normal vector. In practice, we set $\epsilon=1mm$.

\subsection{Training}
\label{sec:training}

For the PeopleSnapshot \cite{alldieck2018detailed} and Multi-Garment Dataset \cite{bhatnagar2019multi}, we use the same hyper parameter across all the subjects. We use Adam \cite{Bengio2015Adam} optimizer for parameter optimization.
Each 3D Gaussian is defined by a center ($x$), scale ($S$), opacity ($\alpha$), rotation ($q$), spherical harmonics ($SH$) and blend weights ($w$). We don't optimize $\alpha$, $q$ and $w$. We keep $\alpha=1$. We optimize the joints position $(J)$ of the canonical skeleton. For learning good geometric details, the number of 3D Gaussians should be enough to model the geometry. We use template mesh of about 220K triangular faces and same number of 3D Gaussians are initialized for PeopleSnapshot and Multi Garment Dataset.

We divide the training into three stages where we prioritize learning geometric information in the earlier stage and color information in the later stages. The first stage lasts till 4\textsuperscript{th} epoch, then the second stage starts from 4\textsuperscript{th} epoch and ends at the 10\textsuperscript{th} epoch and the final stage starts from 11\textsuperscript{th} and lasts till the 20\textsuperscript{th} epoch.
We use the same scale learning rate of 5e-3 and SH encoder learning rate of 5e-4 throughout the training process. We use a joints learning rate of 5e-4 for the first and second stage and do not optimize the joints location in the final stage.
We use learning rate for displacment encoder of 1e-4 in the first and final stage and 8e-4 in the second stage. We keep the normal and photometric loss weight of 1 and normal consistency loss weight of 0.01 throughout the training process. For UBC-Fashion Dataset, we add a pose optimization step where we also optimize the body pose, global orientation and translation parameters of the SMPL body model by using a common learning rate of 1e-4 for all three parameters. 

To obtain the ground truth normal, we use pretrained pix2pixHD network by PIFuHD\cite{saito2020pifuhd}. In Nvidia RTX 4090, normal maps can be obtained for 20 images in less than a second. So, getting normal-map doesn't add any significant time bottleneck in data pre-processing step.

\section{Additional Qualitative Results}
\label{sec:additional_qualitative_results}

\begin{figure}[h]
    \centering
    \includegraphics[width=1\linewidth]{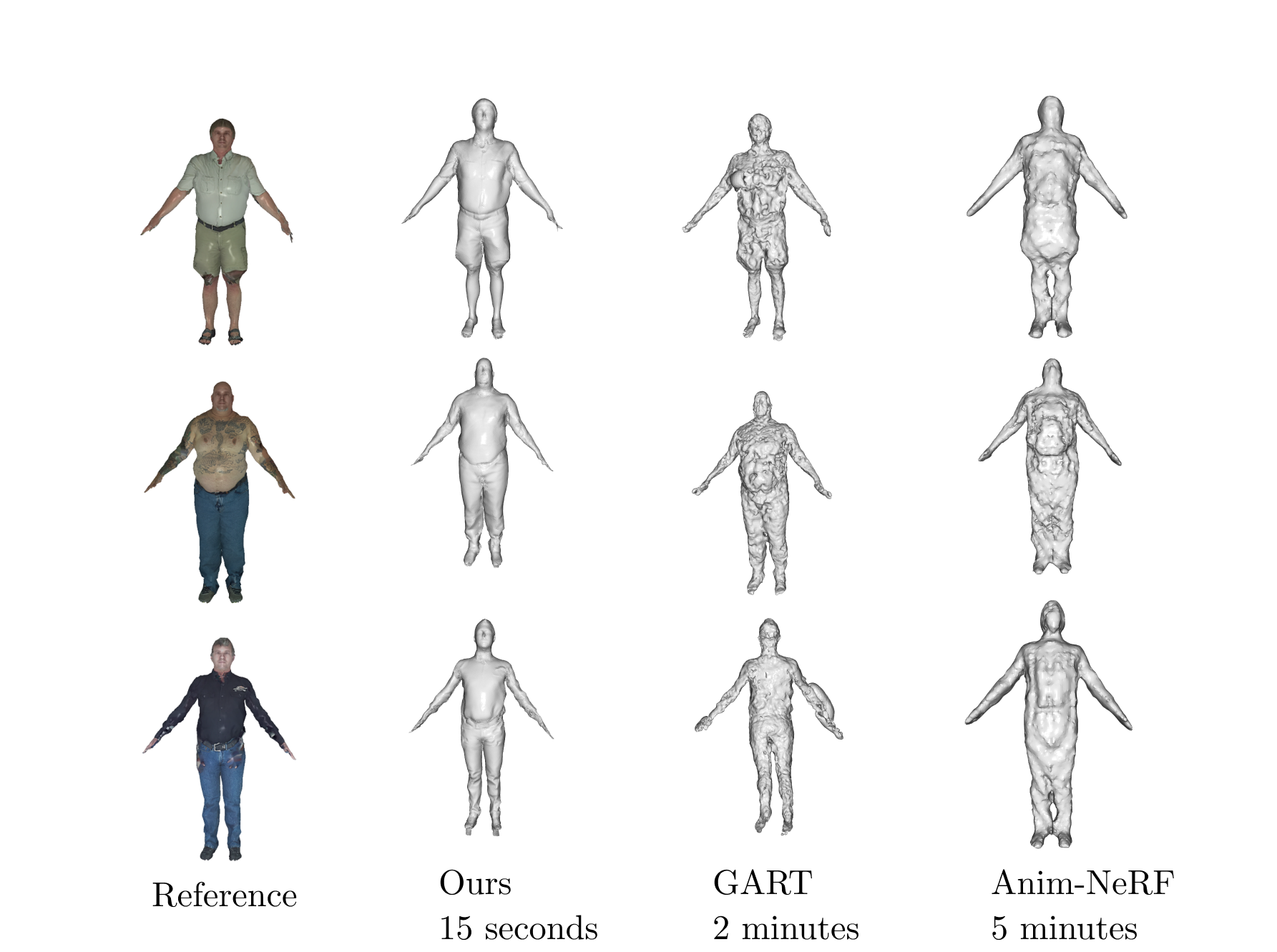}
    \caption{\textbf{Visualization of 3D reconstruction on Multi Garment.} Mesh from GART \cite{lei_gart_2023} and Anim-NeRF \cite{chen2021animatable} suffers from heavy artifacts while our method produces high fidelity mesh. Better surface reconstruction than another 3D Gaussian based method, GART can be attributed to our mesh binding and explicitly computed normal map optimization.}
    \label{fig:mg_3d_comparison}
\end{figure}

\begin{figure}[h]
    \centering
    \includegraphics[width=1\linewidth]{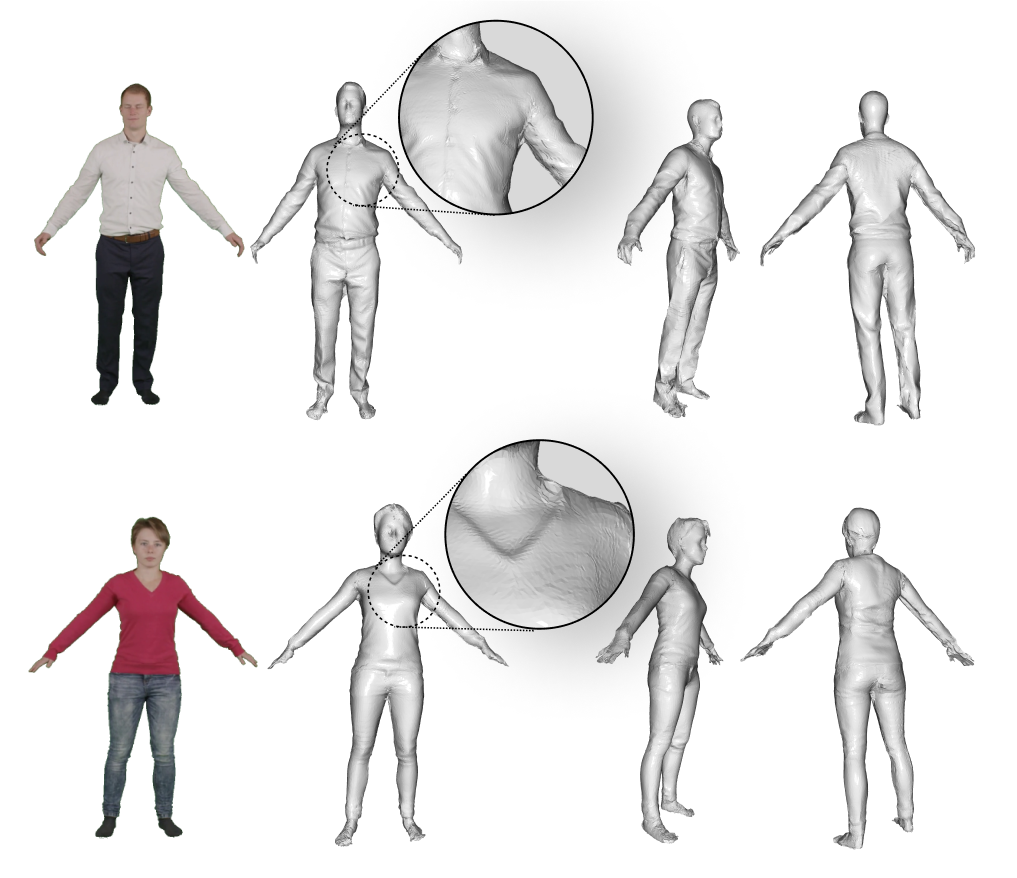}
    \caption{\textbf{More 3D reconstruction results of our method on PeopleSnapshot \cite{DBLP:conf/cvpr/AlldieckMXTP18}}. Our method accurately reconstructs surface details like shirt collar and cloths wrinkles.}
    \label{fig:more_3d_recon_on_peoplesnapshot}
\end{figure}

\textbf{3D Mesh Reconstruction.}
We show 3D mesh reconstruction of GART \cite{lei_gart_2023}, Anim-NeRF \cite{chen2021animatable} and our method on our synthetic Multi-Garment Dataset in Fig. \ref{fig:mg_3d_comparison}. We show more qualitative results of mesh reconstruction of the proposed method on PeopleSnapshot in Fig. \ref{fig:more_3d_recon_on_peoplesnapshot}.

\begin{figure}[h]
    \centering
    \includegraphics[width=1\linewidth]{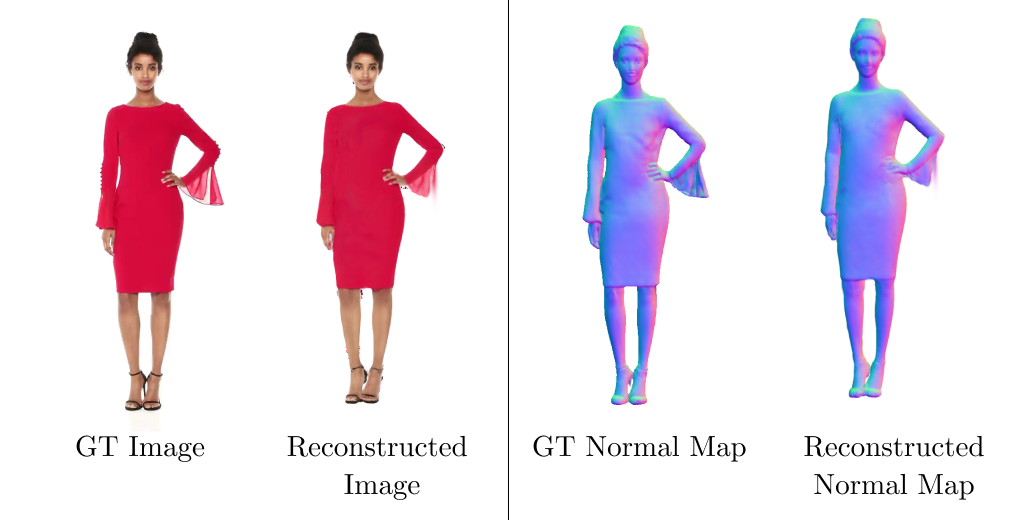}
    \caption{\textbf{View Reconstruction and Normal Map Visualization on UBC-Fashion \cite{zablotskaia2019dwnet}.} (Left) View reconstruction on one subject of UBC-Fashion. (Right) GT normal map and reconstructed normal map visualization.}
    \label{fig:ubc_normal_map}
\end{figure}

In Fig. \ref{fig:ubc_normal_map}, we show reconstructed normal map image. The quality of normal map image being close to the ground truth normal map also shows good quality and fidelity of the reconstructed mesh.

\begin{figure}[h]
    \centering
    \includegraphics[width=1\linewidth]{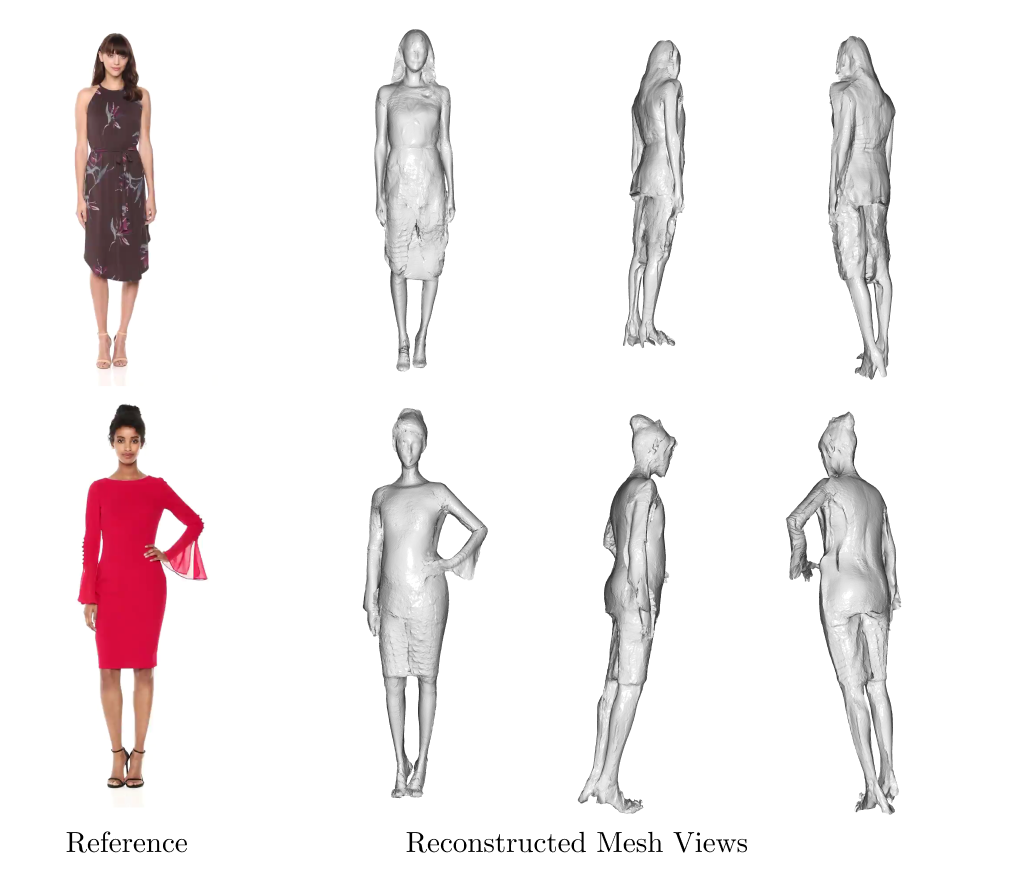}
    \caption{3D Reconstruction on UBC-Fashion \cite{zablotskaia2019dwnet}.}
    \label{fig:ubc_3d_recon}
\end{figure}

We show some qualitative results of 3D Mesh reconstruction on UBC-Fashion dataset in Fig. \ref{fig:ubc_3d_recon}.

\begin{figure}[h]
    \centering
    \includegraphics[width=1\linewidth]{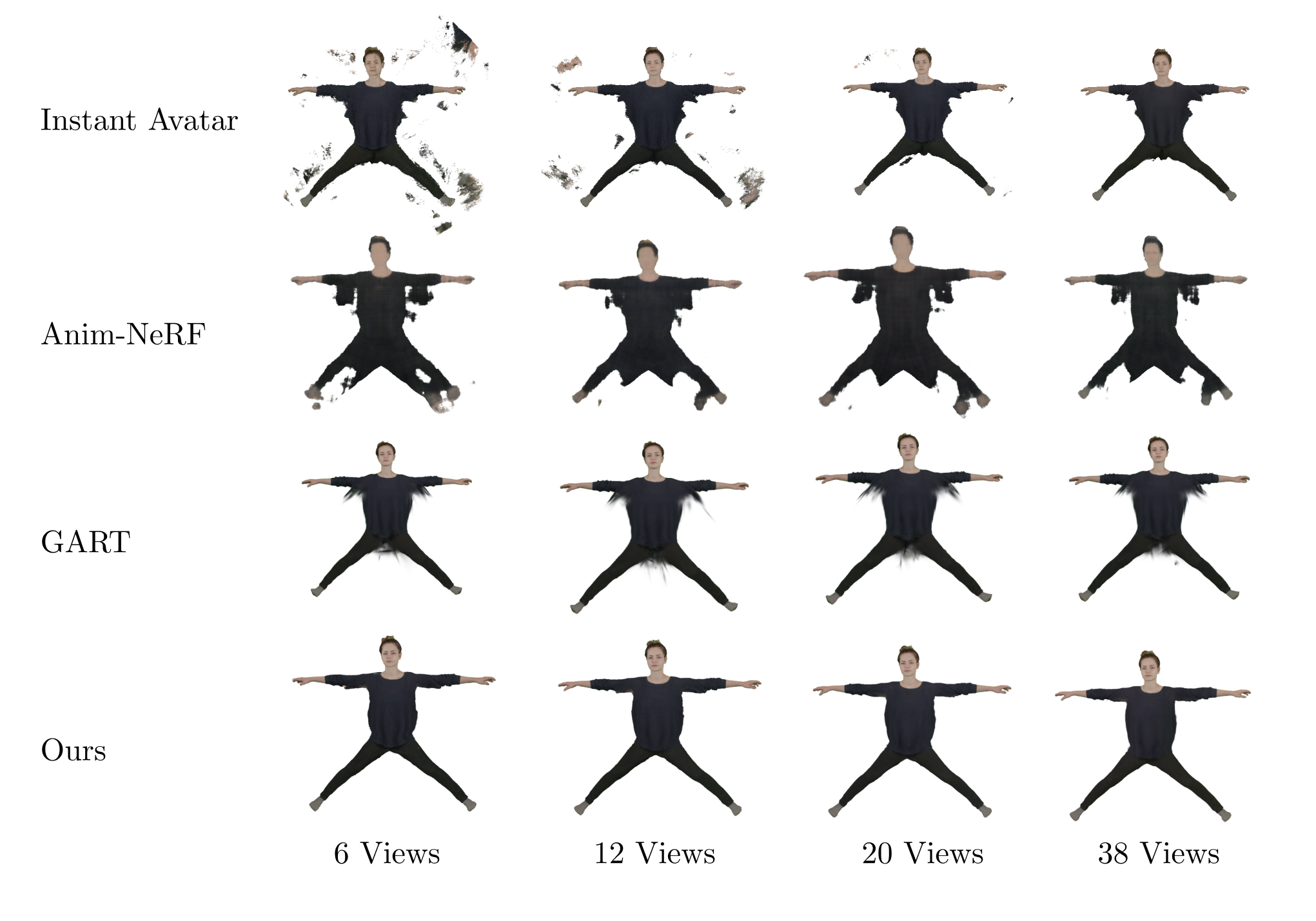}
    \caption{\textbf{Qualitative Comparison of performance of Instant Avatar \cite{jiang2022instantavatar}, Anim-NeRF \cite{chen2021animatable}, GART \cite{lei_gart_2023} and Ours for different number of views in X pose.} Instant Avatar and Anim-NeRF fails to reconstruct the subject whereas GART has artifacts around the arms and between legs. Our method is robust even for only 6 input sequences.}
     \label{fig:comparison_on_num_views_novel_view}
\end{figure}

\textbf{Novel View Synthesis.}
Compared to other SoTA methods, our method achieves novel view with less artifacts in less time and less number of input sequences \ref{fig:comparison_on_num_views_novel_view}.

\begin{figure}[h]
    \centering
    \includegraphics[width=1\linewidth]{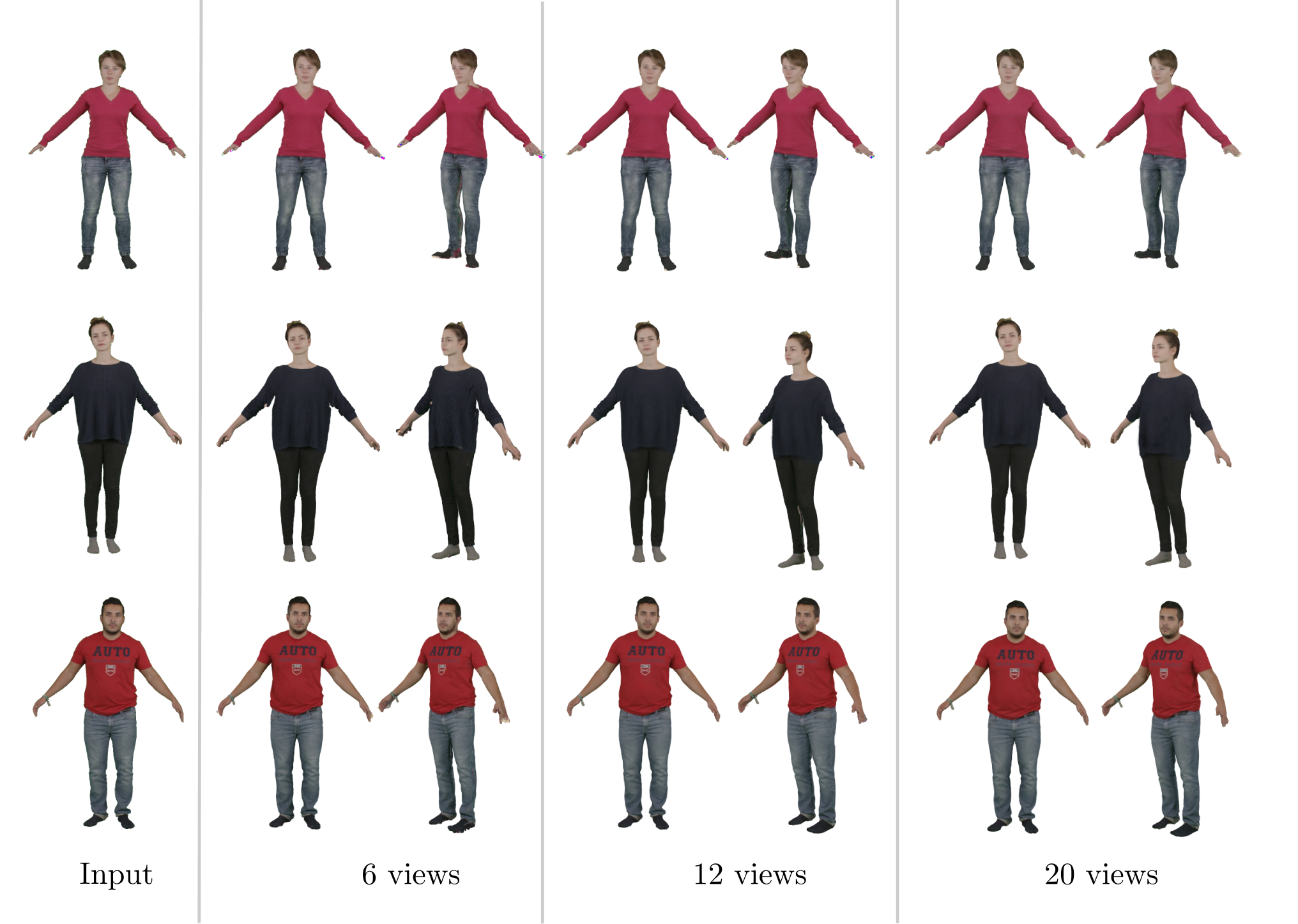}
    \caption{\textbf{Qualitative results on View Synthesis on People Snapshot\cite{alldieck2018detailed} on different number of input sequences.} Even with only 6 input views, our method achieves good quality view reconstruction.}
    \label{fig:novel_view_ours_diff_num_views}
\end{figure}

As shown in Fig. \ref{fig:novel_view_ours_diff_num_views}, iHuman achieves good quality novel view synthesis even being trained with only 6 number of views. 

\begin{figure}[h]
    \centering
    \includegraphics[width=1\linewidth]{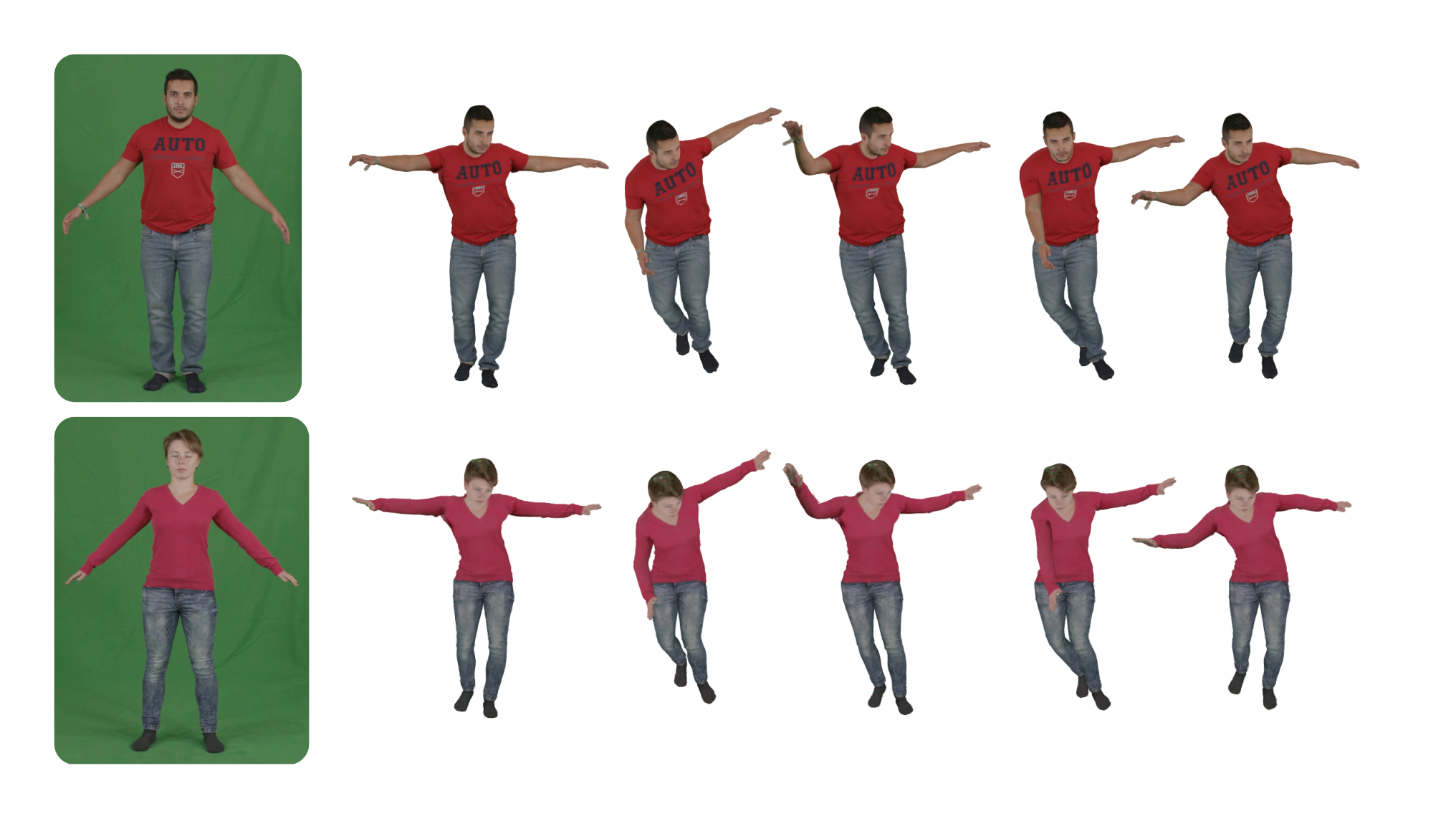}
    \caption{\textbf{Novel Pose Synthesis on PeopleSnapshot \cite{alldieck2018detailed}.} We can feed the SMPL input pose to our iHuman Template Model to synthesize novel poses. Though trained on only A-pose, our method stably renders new image under complex poses.}
    \label{fig:pose_synthesis}
\end{figure}

We show novel pose synthesis results in Fig. \ref{fig:pose_synthesis} on PeopleSnapshot \cite{alldieck2018detailed} of the subjects trained with only 20 input views.

\section{Challenging Cases}
\label{sec:failure_cases}
\begin{figure}
    \centering
    \includegraphics[width=1\linewidth]{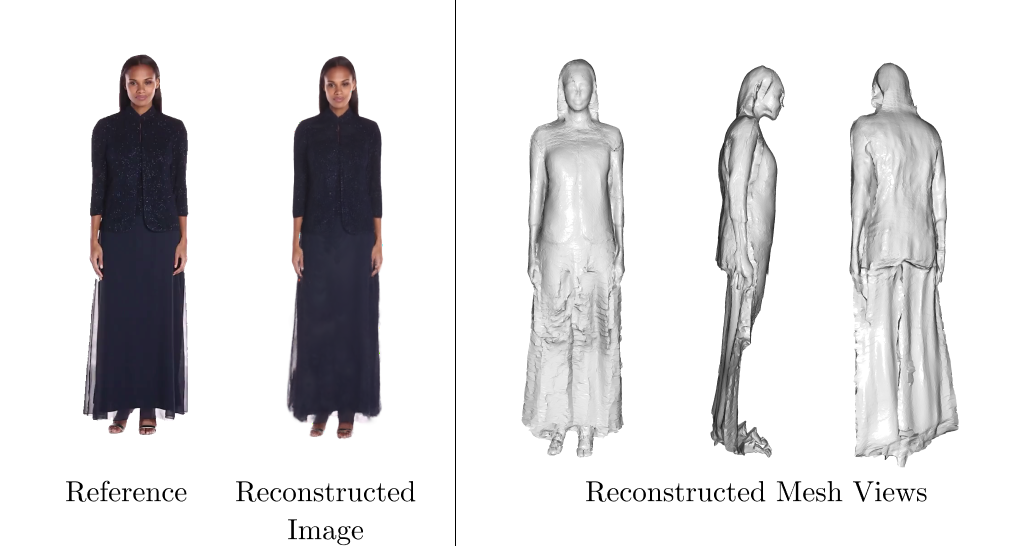}
    \caption{\textbf{Challenging Case.} Under heavy clothing with deformation, the view reconstruction is shown (Left). 3D Mesh reconstruction is implausible for some view direction (Right).}
    \label{fig:challenging_case_ubc}
\end{figure}

UBC-Fashion dataset \cite{zablotskaia2019dwnet} contains subjects in long clothing that undergoes deformation. As shown in Fig. \ref{fig:challenging_case_ubc}, with heavy clothing deformation, though the reconstructed view looks good, there are some geometric implausible views. Even with heavy deforming scene, the reconstructed mesh doesn't contain floating artifacts.

%
%

\clearpage
\bibliographystyle{splncs04}
\bibliography{main}